\documentclass{article}
\usepackage{amsmath}

\usepackage{arxiv}

\usepackage[utf8]{inputenc} % allow utf-8 input
\usepackage[T1]{fontenc}    % use 8-bit T1 fonts
\usepackage{hyperref}       % hyperlinks
\usepackage{url}            % simple URL typesetting
\usepackage{booktabs}       % professional-quality tables
\usepackage{amsfonts}       % blackboard math symbols
\usepackage{nicefrac}       % compact symbols for 1/2, etc.
\usepackage{microtype}      % microtypography
\usepackage{cleveref}       % smart cross-referencing
\usepackage{lipsum}         % Can be removed after putting your text content
\usepackage{graphicx}
\usepackage{doi}
\usepackage[numbers]{natbib}
\usepackage{multirow}
\usepackage{subcaption} % For creating subfigures

\usepackage{amssymb}
\usepackage{mathtools}

\input{defs.tex}
\usepackage{parskip}

\newcommand\blfootnote[1]{%
  \begingroup
  \renewcommand\thefootnote{}\footnote{#1}%
  \addtocounter{footnote}{-1}%
  \endgroup
}
\newcommand\numberthis{\addtocounter{equation}{1}\tag{\theequation}}

\author{ Abhimanyu Das
	\And
	Weihao kong 
    \And
    Rajat Sen
    \And
    Yichen Zhou \\
    \And
    Google Research \\
	\texttt{\{abhidas, weihaokong, senrajat, yichenzhou\}@google.com} \\
}

\title{A decoder-only foundation model for time-series forecasting}

\begin{document}
\maketitle

\begin{abstract}
\blfootnote{Author names are listed in alphabetical order.}
Motivated by recent advances in large language models for Natural Language Processing (NLP), we design a time-series foundation model for forecasting whose out-of-the-box zero-shot performance on a variety of public datasets comes close to the accuracy of state-of-the-art supervised forecasting models for each individual dataset. Our model is based on pretraining a decoder style attention model with input patching, using a large time-series corpus comprising both real-world and synthetic datasets. Experiments on a diverse set of previously unseen forecasting datasets suggests that the model can yield accurate zero-shot forecasts across different domains, forecasting horizons and temporal granularities.

\end{abstract}

\section{Introduction}
\label{sec:intro}

Time-series data is ubiquitous in various domains such as retail, finance, manufacturing, healthcare and natural sciences. In many of these domains, one of the most important use-cases of time-series data is forecasting. Time-series forecasting is critical to several scientific and industrial applications, like retail supply chain optimization, energy and traffic prediction, and weather forecasting. In recent times, Deep learning models~\citep{salinas2020deepar,oreshkin2019n} have emerged as a popular approach
for forecasting rich, multivariate, time-series data, often outperforming classical statistical approaches such as ARIMA or GARCH~\citep{box1968some}. In several forecasting competitions such as the M5 competition~\citep{makridakis2022m5} and IARAI Traffic4cast contest~\citep{pmlr-v133-kopp21a} deep network based solutions performed very well.

At the same time, we are witnessing a rapid progress in the Natural Language Processing (NLP) domain on large foundation models for downstream NLP tasks. Large language models (LLMs) are growing in popularity because they can be used to generate text, translate languages, write different kinds of creative content, and answer your questions in an informative way~\citep{radford2019language}. They are trained on massive amounts of data, which allows them to learn the patterns of human language. This makes them very powerful tools that can be used for a variety of downstream tasks, often in a zero-shot learning mode. 

This motivates the question: “Can large pretrained models trained on massive amounts of time-series data learn temporal patterns that can be useful for time-series forecasting on previously unseen datasets?” In particular, can we design a time-series foundation model that obtains good zero-shot out-of-the-box forecasting performance ? Such a pretrained time-series foundation model, if possible, would bring significant benefits for downstream forecasting users in terms of no additional training burden and significantly reduced compute requirements. It is not immediately obvious that such a foundation model for time-series forecasting is possible. Unlike in NLP, there is no well defined vocabulary or grammar for time-series. Additionally, such a model would need to support forecasting with varying history lengths (context) , prediction lengths (horizon) and time granularities. Furthermore, unlike the huge volume of public text data for pretraining language models, vast amounts of time-series data is not readily available. In spite of these issues, we provide evidence to answer the above question in the affirmative. 

In particular, we design \emph{\ours}, a single foundation model for time-series forecasting that, when applied to a variety of previously-unseen forecasting datasets across different domains, obtains close to state-of-the-art zero-shot accuracy (compared to the best supervised models trained individually for these datasets). Our model can work well across different forecasting history lengths, prediction lengths and time granularities at inference time. The key elements of our foundation model are twofold: 1) a large-scale time-series corpus built using both real-world (mostly time-series data from web search queries\footnote{\url{https://trends.google.com}} and Wikipedia page visits\footnote{\url{https://wikimedia.org/api/rest_v1/}}) and synthetic data, which meets the volume and diversity of data needed for training our foundation model, and 2) a decoder style attention architecture with input patching, that can be efficiently pre-trained on this time-series corpus.

Compared to the latest large language models, our time-series foundation model is much smaller in both parameter size (200M parameters) and pretraining data size (O(100B) timepoints); yet we show that even at such scales, it is possible to pretrain a practical foundation model for forecasting whose zero-shot performance comes close to the accuracy of fully-supervised approaches on a diverse set of time-series data. Our work also suggests that unlike recent work~\citep{gruver2023large} that recommends Large Language Models such as GPT-3 and LLama-2 as out-of-the-box zero-shot forecasters, foundation models trained from scratch exclusively on time-series data can obtain much better zero-shot performance at a tiny fraction of its costs.

\section{Related Work}
\label{sec:related} 
In the last decade, deep learning models~\citep{salinas2020deepar, oreshkin2019n} have emerged as powerful contenders in forecasting time-series in the presence of large training datasets and have been shown to outperform traditional statistical methods such as ARIMA and  Exponential smoothing~\citep{mckenzie1984general}. Forecasting models can be categorized broadly into: (i) Local univariate models that include traditional methods like ARIMA, exponential smoothing~\citep{mckenzie1984general} and non-autoregressive models like Prophet~\citep{taylor2018forecasting}. These models are trained individually for each time-series in a dataset in order to predict the corresponding time-series's future. (ii) Global univariate models like DeepAR~\citep{salinas2020deepar}, Temporal Convolutions~\citep{borovykh2017conditional}, N-BEATS~\citep{oreshkin2019n} and long-term forecasting models such as~\citep{nie2022time, das2023longterm} that are trained globally on many time-series but during inference they predict the future of a time-series as a function of its own past and other related covariates. (iii) Global multivariate models that take in the past of all time-series in the dataset to predict the future of all the time-series. Such models include the classical VAR model~\citep{zivot2006vector} as well as deep learning models like~\citep{sen2019think, zhou2022fedformer, chen2023tsmixer} to name a few.

All the works cited above have primarily been applied in the supervised setting with the notable exception of PatchTST~\citep{nie2022time} and N-BEATS~\citep{oreshkin2019n}. PatchTST has a section on dataset-to-dataset transfer learning in the semi-supervised setting. 
%The input patching in our decoder-only model is inspired by~\citep{nie2022time}.
\citep{oreshkin2021meta} also show that the N-BEATS architecture lends itself to transfer learn between various source-target dataset pairs. However, none of these works aim to train a single foundation model that can work on a plethora of datasets. For an in-depth discussion about transfer learning in time-series we refer the reader to the survey in~\citep{ma2023survey}.

There has been some very recent work on re-using or fine-tuning large language models for time-series forecasting. In particular, ~\citep{gruver2023large} benchmarks pretrained LLMs like GPT-3 and LLaMA-2 for zero-shot forecasting performance. As we show later, our model obtains much superior zero-shot performance at a tiny fraction of these model sizes.
\citep{zhou2023one} and \citep{chang2023llm4ts} show how to fine-tune a GPT-2~\citep{radford2019language} backbone model for time-series forecasting tasks. 
%\citep{chang2023llm4ts} is a follow up works along the same lines.  
%Both the works have a section on zero-shot forecasting on a target dataset after having trained on a source dataset. 
%For instance Table-18~\citep{zhou2023one} shows M4 to M3 transfer.
With the exception of a transfer-learning study (forecasting on a target dataset after having trained on a source dataset), these papers
mostly focus on fine-tuning a pretrained model on target datasets,  and not on pretraining a single foundation model with good out-of-the box zero-shot performance on a variety of datasets. To the best of our knowledge, the very recent work in TimeGPT-1~\citep{garza2023timegpt} is the only other parallel work on a zero-shot foundation model for time-series forecasting. However the model is not public access, and several model details and the benchmark dataset have not been revealed.

% On the other hand, the discussion on a transformer based, patched-decoder style, pre-trained, foundation forecasting model is minimal. Regarding transformers,~\citep{zeng2023transformers} challenged the effectiveness of transformer variants~\citep{wen2022transformers} (\yzedit{}{notably, Informer, Autoformer, FEDformer, etc (cite?)}) on long term time-series forecasting by showing that a simple linear structured model (DLinear) is capable of outperforming those deep, transformer based models. Similarly \citep{oreshkin2019n} introduced N-BEATS, a feedforward model observing block structures and skip connection without attention mechanism that achieved good results. 

% \citep{nie2022time} later justified the use of transformers by involving sub-series level patches and channel independence which achieved the SotA results. It also demonstrated that such model entitled as PatchTST can be pre-trained to later adapt to new domains through finetuning. This pre-training idea is also explored by~\citep{zhou2023one}, in which the authors shared the discovery that by employing the GPT-2 backbone~\citep{radford2019language} and finetuning a task specific head, the resulting model is capable of performing long term time-series forecasting decently well.

% \yzedit{}{tIDE?}

\section{Problem Definition}
\label{sec:pdef}
The task at hand is to build a general purpose zero-shot forecaster that takes in the past $C$ time-points of a time-series as context and predicts the future $H$ time-points. Let the context be denoted by $\*y_{1:L} := \{y_1, \cdots, y_L\}$ where we follow a numpy-like notation for indices. Similarly the actual values in the horizon are denoted by $\*y_{L+1:L+H}$. Note that since we are building a single pre-trained model, we cannot have dataset specific dynamic or static covariates during training time. 
% However, the datetime column is ubiquitous in all time-series data, so we can optionally have date derived features like day or the week, month of the year etc processed into a vector at each time-point $t$, denoted by $\*x_{t} \in \-R^{r}$. 
% %See Appendix~\ref{app:sec:datefeatures} for details. 
% Such features could be available for forecasting in both the context and horizon, represented as $\*x_{1:L+H}$. 
The task is then to learn a foundation model that can map any time-series context to horizon,
\begin{equation}
     f: \left( \*y_{1:L} \right) \longrightarrow \hat{\*y}_{L+1:L+H}.
\end{equation}

The accuracy of the prediction can be measured by a metric that quantifies their closeness to the actual values, for instance, Mean Absolute Error (MAE) defined in Equation~\ref{eq:mae}.

\section{Model Architecture}
\label{sec:model}
A foundation model for time-series forecasting should be able to adapt to variable context and horizon lengths, while having enough capacity to encode all patterns from a large pretraining datasets. Transformers have been shown to be able to adapt to different context lengths in NLP~\citep{radford2019language}. However, there are several time-series specific design choices. The main guiding principles for our architecture are the following:

{\it Patching.} Inspired by the success of patch based modeling in the recent long horizon forecasting work~\citep{nie2022time} we also choose to break down the time-series into patches during training. A patch of a time-series is a natural analogue for a token in language models and patching has been shown to improve performance. Moreover this improves inference speed as the number of tokens being fed into the transformer is reduced by a factor of the patch length. On the other hand, increasing the patch length all the way to the context length moves us away from decoder-only training and the efficiencies that come with it. We delve into this further in Section~\ref{sec:ablation}.

{\it Decoder-only model.} A key difference between our architecture and PatchTST~\citep{nie2022time} is that our model is trained in decoder-only mode~\citep{liu2018generating}. In other words, given a sequence of input patches, the model is optimized to predict the next patch as a function of all past patches. Similar to LLMs this can be done in parallel over the entire context window, and automatically enables the model to predict the future after having seen varying number of input patches.

{\it Longer output patches.} In LLMs the output is always generated in an auto-regressive fashion one token at a time. However, in long-horizon forecasting it has been observed that directly predicting the full horizon yields better accuracy than multi-step auto-regressive decoding~\citep{zeng2022transformers}. But this is not possible when the horizon length is not known apriori, as in the case of zero-shot forecasting which is our primary goal. 

We propose a middle ground by allowing our output patches for prediction to be longer than the input patches. As an example, suppose the input patch length is 32 and output patch length is 128. During training, the model is simultaneously trained to use the first 32 time-points to forecast the next 128 time-steps, the first 64 time-points to forecast time-steps 65 to 192, the first 96 time-points to forecast time-steps 97 to 224 and so on. During inference, suppose the model is given a new time-series of length 256 and tasked with forecasting the next 256 time-steps into the future. The model will first generate the future predictions for time-steps 257 to 384, then condition on the initial 256 length input plus the generated output to generate time-steps 385 to 512. On the other hand, if in a model the output patch length was fixed to the input patch length of 32, then for the same task we would have to go through 8 auto-regressive generation steps instead of just the 2 above. However, there is a trade-off. If the output patch length is too long, then it is difficult to handle time-series whose lengths are less than the output patch length for instance monthly, yearly time-series in our pretraining data.

{\it Patch Masking.} If we use patches naively, the model might only learn to predict well for context lengths that are multiples of the input patch length. Therefore we make a careful use of masking during training. Parts of patches as well as entire patches from the beginning of the context window can be masked in a data batch. We employ a specific random masking strategy (described later) during training that helps the model see all possible context lengths starting from 1 to a maximum context length.

\begin{figure*}[h]
\centering
\includegraphics[width=0.75\linewidth]{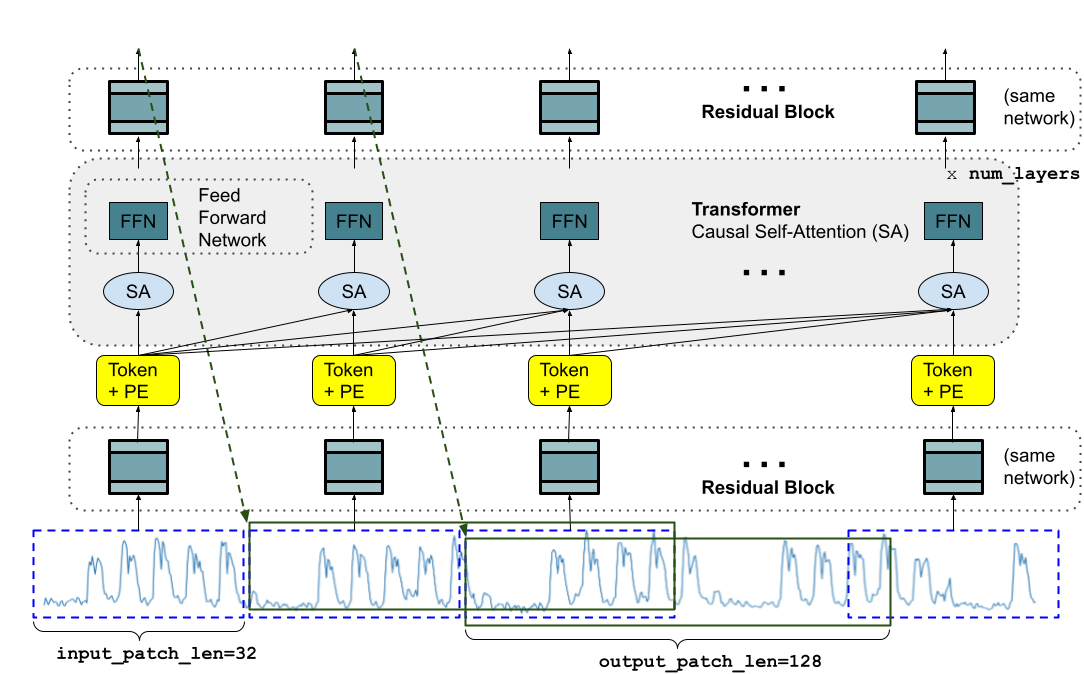}
\caption{We provide an illustration of the \ours~model architecture during training, where we show a input time-series of a specific length that can be broken down into input patches. Each patch along is processed into a vector by a residual block (as defined in the model definition) to the model dimension of the transformer layers. The vector is then added to positional encodings and fed into $n_l$ stacked transformer layers. SA refers to self-attention (note that we use multi-head causal attention) and FFN is the fully connected layer in the transformer. The output tokens are then mapped through a residual block to an output of size \texttt{output\textunderscore patch\textunderscore len}, which is the forecast for the time window following the last input patch seen by the model so far.}
\label{fig:model}
\end{figure*}

Now that we have mentioned the guiding principles,  we next formally describe each component of our model architecture (illustrated in Figure~\ref{fig:model}), which we name as \ours  ~(\textbf{T}ime-\textbf{s}eries \textbf{F}oundation \textbf{M}odel).

{\bf Input Layers.}  The job of the input layers is to preprocess the time-series into input tokens to the transformer layers. We first break the input into contiguous non-overlapping patches. Then each patch is processed by a Residual Block into a vector of size \texttt{model\textunderscore dim}. Along with the input, we also supply a binary padding mask $\*m_{1:L}$ where $1$ denotes that the corresponding input in $\*y_{1:L}$ should be ignored and vice-versa. The Residual Block is essentially a  Multi-layer Perceptron (MLP) block with one hidden layer with a skip connection, similar to that defined in~\citep{das2023longterm}.

% The mixture of experts implementation we use is identical to the top-2 gating using in~\citep{lepikhin2020gshard}. We shall see in Section~\ref{sec:emp} that the MoE layer helps the network adapt to different granularities. 

In other words, the inputs $\*y_{1:L}$ are broken down into patches of size \texttt{input\textunderscore patch\textunderscore len} ($p$). The $j$-th patch can be denoted as  $\tilde{\*y}_j = \*y_{p(j-1)+1:pj}$.  Similarly the mask can also be patched as $\tilde{\*m}_j = \*m_{p(j-1)+1:pj}$. Then the $j$-th input token to the subsequent transformer layers can be denoted as,
\begin{align}
    \*t_j = \texttt{InputResidualBlock}(\tilde{\*y}_j\odot (1 - \tilde{\*m}_j)) + \texttt{PE}_j \label{eq:input}
\end{align}
where $\texttt{PE}_j$ denotes the $j$-th positional encoding as defined in the original transformer paper~\citep{vaswani2017attention}. There will be $N = \floor{L/p}$ such input tokens.

{\bf Stacked Transformer.} The bulk of the parameters in our model are in \texttt{num\textunderscore layers} ($n_l$) transformer layers stacked on top of each other. Each of these layers have the standard multi-head self-attention (SA) followed by a feed-forward network (FFN). The main hyperparameters are \texttt{model\textunderscore dim} which is equal to the dimension of the input tokens $\*t_j$'s and number of heads (\texttt{num\textunderscore heads}). We set the hidden size of the FFNs to be equal to \texttt{model\textunderscore dim} as well. We use causal attention that is each output token can only attend to input tokens that come before it in the sequence (including the corresponding input token). This can be described by the equation
\begin{align*}
    \*o_j = \texttt{StackedTransformer} ( (\*t_1,\dot{m}_1), & \cdots, (\*t_{j},  \dot{m}_j ) ), \numberthis \label{eq:stacked}
\end{align*}
for all $j \in [N]$. $\dot{m}_j$ is the masking indicator for the $j$-th token defined as $\min\{\*m_{p(j-1)+1:pj}\}$ i.e if a patch has any non-masked time-point the corresponding token marked as not being masked. All patches that are masked out completely are not attended to by the causal self attention.

% \begin{remark}
% \label{rem:padding}
% It might seem that with the above architecture we can only support context lengths that are multiples of \texttt{input\textunderscore patch\textunderscore len} ($p$). However, with a simple use of padding we can support any context length ranging from 1 to maximum training context length. We supply a binary padding mask $\*m_{1:L}$ where $1$ denotes that the corresponding input in $\*x_{1:L}$ should be ignored and vice-versa. Then $\tilde{\*x}_j$ can be pointwise multiplied by $1 - \*m_{p(j-1):pj}$ before feeding it to Eq.~\eqref{eq:input}. Moreover patches that are completely masked out can be excluded from the attention in Eq.~\eqref{eq:stacked}.
% \end{remark}

{\bf Output Layers.} The remaining task is to map the output tokens into predictions. We train in decoder only mode i.e each output token should be able to be predictive of the part of the time-series that follows the last input patch corresponding to it. This is common for popular large language models like~\citep{radford2019language}. However, one key difference in our time-series foundation model is that input patch length need not be equal to output patch length i.e we should be able to predict a larger chunk of the time-series based on the encoded information from the input patches seen so far. Let the output patch length be \texttt{output\textunderscore patch\textunderscore len} ($h$). We use another Residual Block to map the output tokens to the predictions. This can be described as,
\begin{align}
    \hat{\*y}_{pj+1: pj + h} = \texttt{OutputResidualBlock} \left( \*o_j \right). \label{eq:output}
\end{align}

Thus we encode all the data in $\*y_{1:pj}$ into $\*o_j$ and use that to predict the subsequent $h$ time-points $\*y_{pj+1: pj + h}$. This is done for all patches in one training mini-batch.

{\bf Loss Function.} In this work, we focus on point forecasting. Therefore we can use a point forecasting loss during training like Mean Squared Error (MSE). The loss that is minimized during training can be expressed as,
\begin{align}
    \texttt{TrainLoss} = \frac{1}{N}\sum_{j=1}^{N} \mathrm{MSE}\left(\hat{\*y}_{pj+1: pj + h}, \*y_{pj+1: pj + h}\right).
\end{align}

Note that if one is interested in probabilistic forecasting, then it is easy to have multiple output heads for each output patch, each head minimizing a separate quantile loss as in~\citep{wen2017multi}. Another approach can be to output the logits of a probability distribution family and minimize the maximum likelihood loss for probabilistic forecasting~\citep{awasthi2021benefits,salinas2020deepar}.

{\bf Training.} We train the model with standard mini-batch gradient descent in decoder-only fashion, that goes through all windows for a time-series and across time-series. The only non-standard part is the way we sample the mask during training. For each time-series in the batch, we sample a random number $r$ between $0$ and $p-1$. Then we set the $\*m_{1:r} = 1$ and the rest as zero i.e we mask \rev{out} a fraction of the first input patch. However, this is sufficient to cover all input context lengths from 1 to the maximum training context length. We explain this using an example below:

Suppose the maximum context length is 512 and $p=32$. Then if $r=4$, the output prediction after seeing the first patch (from $\*o_1$) is optimized to predict after seeing $28=32-4$ time-points, the output of the next patch (from $\*o_2$) is optimized to predict after seeing $28+32$ time-points, and so on. When this argument is repeated for all such $r$'s, the model has seen all possible context lengths till $512$.

{\bf Inference.} The trained network can be used to produce forecasts for \textit{any} horizon using auto-regressive decoding similar to large language models. Given an input $\*y_{1:L}$ (assume $L$ is a multiple of $p$ for simplicity) it can first predict  $\hat{\*y}_{L+1: L + h}$. Then, we can use the concatenated vector $\tilde{\*y}_{1:L+h} = [\*y_{1:L}; \hat{\*y}_{L+1: L + h}]$ as an input to the network to generate the next output patch prediction $\hat{\*y}_{L+h+1: L + 2h}$ and so on. If $L$ is not a multiple of $p$, we simply append  zeros to make it a multiple of $p$ and mark the corresponding entries in the mask as 1.

%We name our model \textbf{Pre}trained \textbf{D}e\textbf{c}oder for \textbf{T}ime-series (\ours).
%We name our \textbf{T}ime-\textbf{s}eries \textbf{F}oundation \textbf{M}odel as \ours.

\section{Pretraining Details} 
\label{sec:pretrain}

We would like our pretraining corpus to include large volumes of temporal data representing a variety of domains, trend and seasonality patterns and time granularities that ideally capture the forecasting use-cases which we are interested in serving by the deployed model. It is challenging to find a large time-series dataset that meets the volume and diversity of data needed for training our foundation model. We address this problem by sourcing the bulk of data used to train our models from three major sources: Google trends, Wiki Pageview statistics and synthetic time-series. In summary the main data sources are:

{\it Google Trends.} Google Trends~\footnote{\url{https://trends.google.com}} captures search interest over time for millions of queries. We choose around~22k head queries based on their search interest over 15 years from 2007 to 2022. Beyond these head queries the time-series become more than 50\% sparse. We download the search interest over time for these queries in hourly, daily, weekly and monthly granularities to form our dataset. The date ranges are Jan. 2018 to Dec. 2019 for hourly and Jan. 2007 to Dec. 2021 for the other granularities.  The trends datasets amounts to roughly \rev{0.5B} time-points.

{\it Wiki Pageviews.} Wiki Pageviews~\footnote{\url{https://en.wikipedia.org/wiki/Wikipedia:Pageview_statistics}} captures the hourly views of all Wikimedia pages. We download all pageview data from Jan. 2012 to Nov. 2023, clean and aggregate the views by page into hourly, daily, weekly and monthly granularities, and filter out pageview time-series with excessive zeros. The final corpus contains roughly \rev{300B} time-points.

{\it Synthetic Data.} Another major component of our pretraining data is of synthetic origin. We create generators for ARMA~\citep{mckenzie1984general} processes, seasonal patterns (mixture of sines and cosines of different frequencies), trends (linear, exponential with a few change-points) and step functions. A synthetic time-series can be an additive combination of one or more of these processes. We create 3M synthetic time-series each of length 2048 time-points. More details about our synthetic data generation are presented in Appendix~\ref{app:syn}.

{\it Other real-world data sources.} Along with the wiki and trends data, we also add time-series from several other publicly available datasets to our pretraining corpus. We add all the granularities of the M4 dataset~\citep{makridakis2022m5}, the hourly and 15 minute Electricity and the hourly Traffic datasets (see ~\citep{zhou2021informer}). We also add the 10-minute granularity Weather dataset used for evaluations in~\citep{zhou2021informer}. M4 has a good mix of granularities with around~100k time-series in total. Traffic and Electricity are  large long-term forecasting datasets with $>$ 800 and $>$ 300 time-series each having tens of thousands of time-points. In addition, we add all the 15 min granularity traffic time-series from~\citep{libcitylong}.

\begin{table}[!ht]
    \centering
    \caption{Composition of TimesFM pretraining dataset.}
    % \resizebox{0.\textwidth}{!}{
    \begin{tabular}{llrr}
    \toprule
    \multicolumn{1}{c}{Dataset} & \multicolumn{1}{c}{Granularity} & \multicolumn{1}{c}{\# Time series} & \multicolumn{1}{c}{\# Time points} \\ \midrule
    Synthetic & & 3,000,000 & 6,144,000,000  \\
    Electricity & Hourly & 321 & 8,443,584  \\
    Traffic & Hourly & 862 & 15,122,928  \\
    Weather~\citep{zhou2021informer} & 10 Min & 42 & 2,213,232  \\
    Favorita Sales & Daily & 111,840 & 139,179,538  \\
    LibCity~\citep{libcitylong} & 15 Min & 6,159 & 34,253,622  \\
    M4 hourly & Hourly & 414 & 353,500  \\
    M4 daily & Daily & 4,227  & 9,964,658  \\
    M4 monthly & Monthly & 48,000 & 10,382,411  \\
    M4 quarterly & Quarterly & 24,000 & 2,214,108  \\
    M4 yearly & Yearly & 22,739 & 840,644  \\
    Wiki hourly & Hourly & 5,608,693 & 239,110,787,496  \\
    Wiki daily & Daily & 68,448,204 & 115,143,501,240  \\
    Wiki weekly & Weekly & 66,579,850 & 16,414,251,948  \\
    Wiki monthly & Monthly & 63,151,306 & 3,789,760,907  \\
    Trends hourly & Hourly & 22,435 & 393,043,680  \\
    Trends daily & Daily & 22,435 & 122,921,365  \\
    Trends weekly & Weekly & 22,435 & 16,585,438  \\
    Trends monthly & Monthly & 22,435 & 3,821,760 \\
    \bottomrule
    \end{tabular}
    % }
    \label{tab:train_data}
\end{table}
{\bf Dataset Mixing and Training.} We train on a mixture distribution over these datasets that aims to give sufficient weight to all granularities \rev{and datasets}. The training loader samples \rev{80\%} real data and \rev{20\%} synthetic, with the real data mixture providing equal weights to the groups: hourly + sub-hourly, daily, weekly, and monthly datasets.
We train with a maximum context length of 512 whenever the length of the time-series allows that. For weekly granularity we do not have sufficiently long time-series; therefore a maximum context length of 256 is used. For the same reason, a maximum context length of 64 is used while training on $\geq$ monthly granularity data. We also use only the standard normalization part of reversible instance normalization~\citep{kim2021reversible} -- i.e, the context of each time-series is scaled by the context mean and standard deviation of the first input patch in the context.

\begin{figure*}[!ht]
  \centering
  \begin{subfigure}[b]{0.3\textwidth}
     \includegraphics[width=\textwidth]{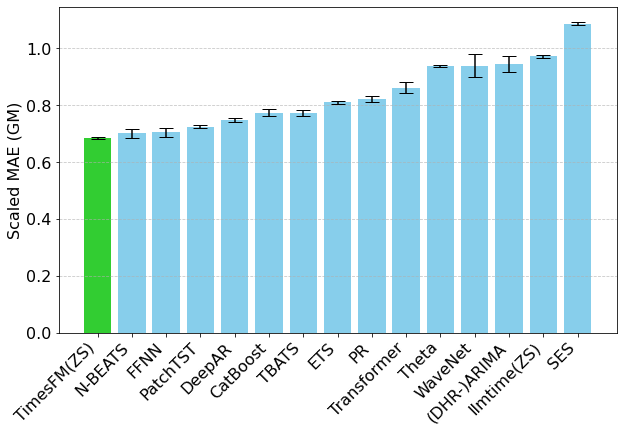} % Replace 'placeholder1' with your image filename
     \caption{\scriptsize{\rev{Monash Archive}~\citep{godahewa2021monash}}}
     \label{fig:monash_mae}
  \end{subfigure}
  \hfill % Add some space between subfigures
  \begin{subfigure}[b]{0.3\textwidth}
     \includegraphics[width=\textwidth]{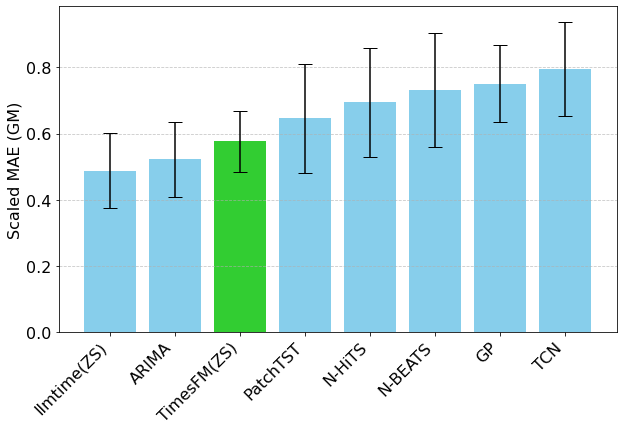} % Replace 'placeholder2' with your image filename
     \caption{\scriptsize{\rev{Darts}~\citep{herzen2022darts}}}
     \label{fig:darts}
  \end{subfigure}
  \hfill % Add some space between subfigures
  \begin{subfigure}[b]{0.3\textwidth}
     \includegraphics[width=\textwidth]{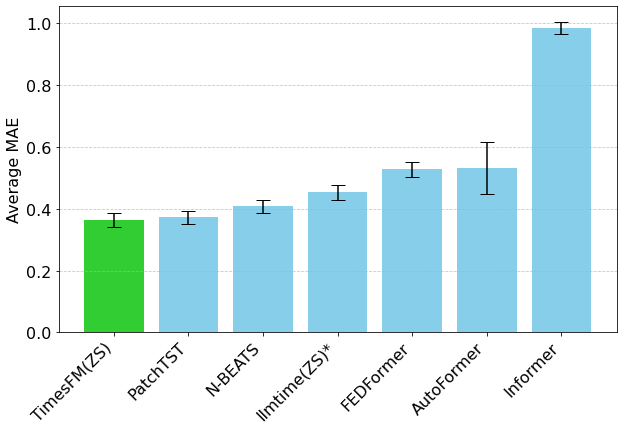} % Replace 'placeholder3' with your image filename
     \caption{\scriptsize{ETT (Horizons 96 and 192)~\citep{zhou2021informer}}}
     \label{fig:informer}
  \end{subfigure}
\caption{We report average performance in three groups of datasets. In all figures, the lower the metric the better and the error bars represent one standard error. Note that among the baselines \rev{only} \ours~and llmtime are zero-shot. In (a) we report results on the Monash datasets. \rev{Since the datasets have different scales, we take the Geometric Mean (GM) of the MAE's scaled by the MAE of a naive baseline}. We can see that \ours~is the top model. In (b), we report the similarly scaled MAE on the Darts benchmarks. \ours~is within significance of the best performing methods which are ARIMA and llmtime in this case. Note that these datasets have one time-series each and therefore statistical methods are competitive with deep learning ones. Finally, in (c) we report the \rev{average} MAE for 96 and 192 horizon prediction tasks on 4 ETT datasets i.e 8 tasks in total. \ours~and PatchTST are the best performing models}
  \label{fig:all_bar_charts}
\end{figure*}

\section{Empirical Results}
\label{sec:emp}

We evaluate our model in zero-shot settings on three groups of well known public datasets against the best performing baselines for each group. These datasets have been intentionally held out from our pretraining data. We show that a \textit{single} pretrained model can come close or surpass the performance of baselines models on the benchmarks even when the baselines are specially trained or tuned for each specific task. Subsequently, we perform ablation studies that justify different choices made in our architecture. 

\subsection{Zero-shot Evaluation}
\label{sec:zeroshot}
To benchmark our model’s performance, we choose three groups of commonly used forecasting
datasets that cover various domains, sizes, granularities, and horizon lengths: Darts~\citep{herzen2022darts}, Monash~\citep{godahewa2021monash} and Informer datasets~\citep{zhou2021informer}, to test the generalization power of our foundation model against other baselines.

In all cases, we report performance on the official metrics and scalings of the datasets, \rev{using either their standard test splits or common test splits in other literature}. We present a summary of the results below - more details can be found in Appendix~\ref{app:moreexp}. We provide the hyper-parameters and other details about our model in Appendix~\ref{app:moremodels}.

% In order to present the average results on each dataset groups we need to aggregate the metrics in such a way that they reflect the hardness of the individual datasets and also not be skewed by different scales in each dataset. Therefore, we report the percentage improvement of the algorithms over a \textit{Naive} method that outputs the constant prediction set to the mean of the context window. The percentage improvements are then averaged across all the datasets in each group.

{\bf Monash~\citep{godahewa2021monash}.} Monash archive is a collection of 30 datasets of different training and prediction lengths that covers granularities ranging from minutes to years and domains including finance, demand forecasting, weather and traffic. The archive reports four official metrics for several statistical baselines such as Exponential Smoothing(ETS) and ARIMA, as well as supervised ML baselines like CatBoost~\citep{prokhorenkova2018catboost}, DeepAR~\citep{salinas2020deepar} and WaveNet~\citep{oord2016wavenet}. Following llmtime~\citep{gruver2023large} we start from the Monash Huggingface repository~\footnote{\url{https://huggingface.co/datasets/monash_tsf}} and filter out the datasets that contain missing values. This leaves us with 18 datasets which we specify in Appendix~\ref{app:monash}.

Out of the four official metrics, following prior work~\citep{gruver2023large}, we report our performance in terms of mean MAE (see Appendix~\ref{app:metrics}). As the datasets have massively different scales, for each dataset we normalize the metric by the metric achieved by a naive baseline that just constantly predicts the last value in the context for each time-series. Then the scaled MAE's are averaged across all datasets. The scaled aggregation was also used in~\citep{gruver2023large}. \rev{In Figure~\ref{fig:monash_mae}, we use the Geometric Mean (GM) for averaging since it is more robust for normalized metrics~\citep{fleming1986not}. We also report the Arithmetic Mean based aggregated metrics in Figure~\ref{fig:all_bar_charts_am} in the appendix.}

The mean scaled MAE across all datasets is plotted in Figure~\ref{fig:monash_mae} along with standard error bars. We compare the performance of \ours~with the baseline models implemented in Monash, and the zero-shot llmtime~\citep{gruver2023large} model that uses GPT-3~\citep{radford2019language} with a specific prompting technique. Note that the zero-shot models are marked as (\textbf{Z}ero-\textbf{S}hot). \rev{\ours~is the top model even though we never trained on these datasets. It is slightly better but within significance of N-BEATS but outperforms deep supervised models like DeepAR~\citep{salinas2020deepar}, and improves on llmtime's performance by more than 25\%}. 

{\bf Darts~\citep{herzen2022darts}.} This is a collection of 8 univariate datasets which include interesting seasonalities and additive+multiplicative trends. We report performance of several baselines implemented in the Darts package like TCN~\citep{lea2016temporal}, N-HiTS~\citep{challu2022nhits} and N-BEATS~\citep{oreshkin2019n}. All these baselines are supervised. As before,  we also report zero-shot forecasting results from llmtime~\citep{gruver2023large} using GPT-3~\citep{radford2019language}. Other supervised baselines in~\citep{gruver2023large} like SM-GP~\citep{wilson2013gaussian} and ARIMA~\citep{mckenzie1984general} are also added. 

We report  the official metric for this dataset group that is MAE for each individual dataset in Appendix~\ref{app:moreexp}. In Figure~\ref{fig:darts}, we present the average scaled MAE across all 8 datasets, as we did for the Monash datasets. \ours~is within statistical significance of the best models that is llmtime and seasonal ARIMA in this case. Note that since there are only 8 individual time-series in this dataset group, the standard errors are not sharp and therefore does not provide a clear ordering among the models. Also, note that for ARIMA, the seasonality needs to be encoded correctly in the parameters for the best results, which needed manual tuning. Further, since these datasets are used in numerous time series blog posts for illustrative purposes, data contamination for llmtime cannot be ruled out.

{\bf Informer~\citep{zhou2021informer}.} The Informer datasets have been widely used for benchmarking various supervised long-horizon forecasting methods. A few of these datasets are used in pretraining, so we focus on the other datasets in this collection (ETTm1, ETTm2. ETTh1 and ETTh2) related to electricity transformer temperatures over a two year period in $1$ hour and $15$ minutes granularities. Note that the long horizon baselines usually report rolling validation results on the test set which would amount to millions of tokens for evaluating llmtime~\citep{gruver2023large} and would be too expensive. Therefore, following llmtime, we compare all methods on the last test window. Also, it is reasonable to directly average the MAE for these datasets since the results are reported on standard normalized dataset (using the statistics of the training portion).

We consider the task of predicting horizon length 96 and 192, given a context length of 512 for all methods. The MAE averaged over all 8 tasks (4 datasets with two horizons each) is presented in Figure~\ref{fig:darts}. \ours~performs the best and the supervised PatchTST~\citep{nie2022time} baseline (which is a state-of-the-art long horizon deep forecasting method) is within significance of it. The other long horizon methods are quite a bit worse even though they have been trained these datasets. llmtime is better than FEDFormer but worse than PatchTST with statistical significance. 

\rev{We present visual examples of our forecasts along with baselines in Appendix~\ref{app:syn_exp}.}

\subsection{Ablation}
\label{sec:ablation}

\begin{figure*}[!ht]
    \centering
    \begin{subfigure}[b]{0.4\textwidth}
       \includegraphics[width=\textwidth]{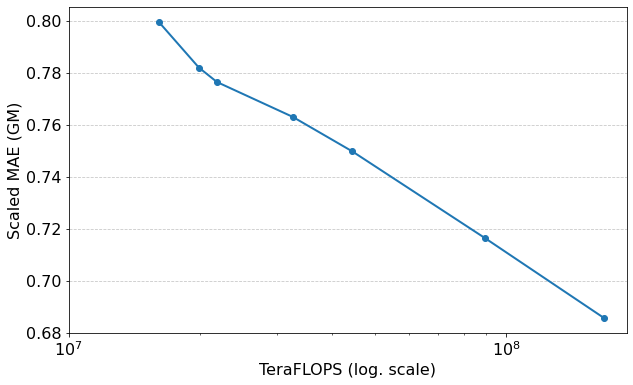} % Replace 'placeholder1' with your image filename
       \caption{Scaled MAE (GM) on Monash datasets as a function of FLOPS across three model sizes 17M, 70M and 200M. The first 4 points are from 17M and 70M checkpoints while the last 3 are from 200M.}
       \label{fig:scale}
    \end{subfigure}
    \hfill % Add some space between subfigures
    \begin{subfigure}[b]{0.4\textwidth}
       \includegraphics[width=\textwidth]{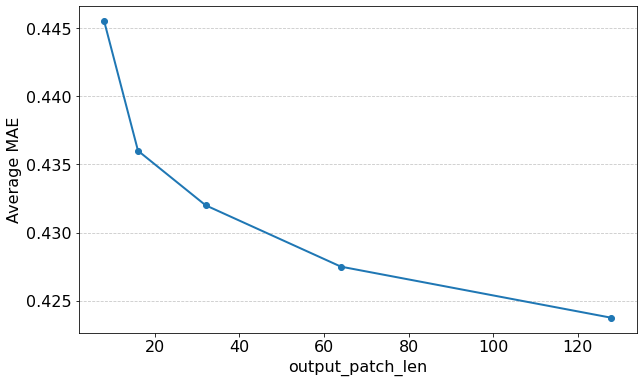} % Replace 'placeholder2' with your image filename
       \caption{Ablation with respect to output patch length for the task of predicting 512 steps into the future on ETT datasets on the original test set in~\citep{zhou2021informer}. We report the average across all 4 ETT datasets.}
       \label{fig:auto}
    \end{subfigure} \\
    \begin{subfigure}[b]{0.4\textwidth}
       \includegraphics[width=\textwidth]{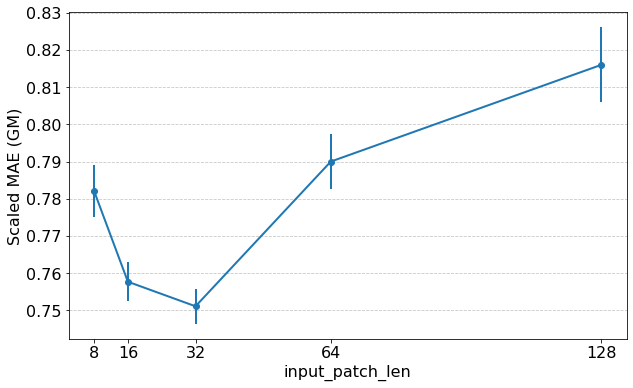} % Replace 'placeholder3' with your image filename
       \caption{Scaled MAE (GM) for our 70M models on Monash datasets for different input patch lengths. We also plot error bars denoting one standard error.}
       \label{fig:iplen}
    \end{subfigure}
    \hfill % Add some space between subfigures
    \begin{subfigure}[b]{0.4\textwidth}
        \includegraphics[width=\textwidth]{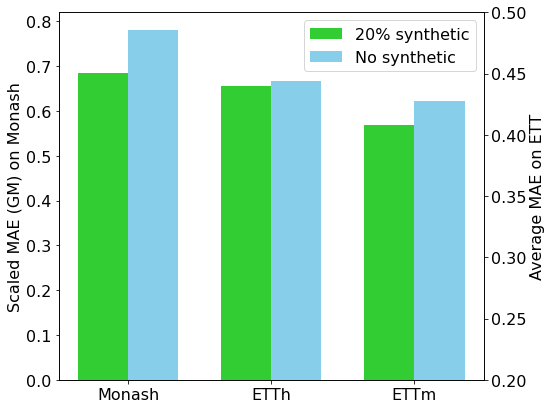} % Replace 'placeholder3' with your image filename
        \caption{Average scaled MAE for Monash on the left and average MAE on ETT datasets on the right. We compare the performance of 200M model with and without the synthetic data.}
        \label{fig:tbbbb}
     \end{subfigure}
  \caption{Ablation studies with respect to various design choices.}
    \label{fig:datasets-abl}
  \end{figure*}

Next, we perform several ablation studies that inform the design decisions we made for our  model architecture.

{\bf Scaling.} Performance curves with respect to number of parameters in a model have been a keenly studied area in the context of LLMs. ~\citep{kaplan2020scaling} established a power law like relationship between the number of parameters in a language model and its downstream performance i.e the more the number of paramaters the better the performance. However, ~\citep{hoffmann2022training} established a more nuanced scaling law that lays down methods to train compute optimal models based on the number of tokens available in a training dataset. 

\rev{We perform a preliminary scaling study where we train three \ours~models of sizes 17M, 70M and 200M parameters, using the same pre-training dataset till 1.5M iterations with a global batch-size of 4096. Then we collect checkpoints that represent varying number of FLOPS (Floating Point OPerationS) across the different model runs. Then we plot the performance on Scaled MAE (GM) on Monash as a function of FLOPS, in Figure~\ref{fig:scale}. This is now a standard way to perform scaling studies in LLMs (see recent work like~\citep{gu2023mamba})}. It can be clearly seen that the errors decrease monotonically with the number of FLOPS (in log scale). \rev{All experiments were performed on a TPUv5e\footnote{\url{https://cloud.google.com/tpu/docs/v5e-training}} setup with 16 tensor-cores. For the 200M model it takes 2 days to complete 1.5M iterations on our setup.}

% \begin{figure}[!ht]
%     \centering
%     \includegraphics[width=0.8\linewidth]{scaling_flops.png}
% \caption{\rev{Scaled MAE (GM) on Monash datasets as a function of FLOPS across three model sizes 17M, 70M and 200M. The first 4 points are from 17M and 70M checkpoints while the last 3 are from 200M.}}
% \label{fig:scale}
% \end{figure}

{\bf Autoregressive Decoding.} In recent long-term forecasting works~\citep{zeng2022transformers,nie2022time,das2023longterm} it has been observed that directly predicting the entire forecasting horizon in one shot  from a decoder can yield better results than auto-regressive decoding on long horizon benchmarks. For a foundation model, the horizon length of the task is not known before inference time, therefore one-shot decoding might not be possible for very long horizons. However, as mentioned earlier, by keeping the \texttt{output\textunderscore patch\textunderscore len} longer than \texttt{input\textunderscore patch\textunderscore len} one can ensure fewer autoregressive steps. This was one of the key decisions in the design of \ours, that is quite different from LLMs. \rev{In order to showcase this we choose the task of predicting 512 time-steps into the future for the ETT datasets on the original rolling validation task of the ETT test sets~\citep{zhou2021informer}. In Figure~\ref{fig:auto}, we present results from models with output\textunderscore patch\textunderscore len varying from 8 to 128. We see a monotonic decrease in average MAE with output\textunderscore patch\textunderscore len.}

% \begin{figure}[!ht]
%     \centering
%     \includegraphics[width=0.7\linewidth]{op_len_new.png}
% \caption{Ablation with respect to output patch length for the task of predicting 512 steps into the future on ETT datasets on the original test set in~\citep{zhou2021informer}. We report the average across all 4 ETT datasets.}
% \label{fig:auto}
% \end{figure}

{\bf Input Patch Length.~} The size of \texttt{input\textunderscore patch\textunderscore len} represents an important trade-off. We have typically seen that increasing its value from 8 to 32 increases performance but having too high a \texttt{input\textunderscore patch\textunderscore len} is impractical since that makes the model shift from decoder only training more towards encoder-decoder style training. Note that in the "Training" paragraph of Section~\ref{sec:model}, we describe the mask sampling strategy to support any context length. If in the extreme case $p$ is set the maximum context length we have to individually sample all possible context windows from 1 to maximum context length, which would be required for encoder-decoder style of training.

\rev{In Figure~\ref{fig:iplen}, we show the mean scaled MAE (GM) \ours(ZS) - 70M model on Monash with \texttt{input\textunderscore patch\textunderscore len} varying from 8 to 128. Note that both models have been trained to about 1.5M steps even though the p=8 model is three times slower to train. We can see that $p=16,32$ marks the best performance, with the error increasing towards either end. Note that $p=32$ model is almost twice as fast to train compared to $p=16$ and thus constitutes a prudent choice.}

% \begin{figure}[!ht]
%     \centering
%     \includegraphics[width=0.7\linewidth]{ip_len_abl_detailed.png}
% \caption{Scaled MAE (GM) for our 70M models on Monash datasets for different input patch lengths. We also plot error bars denoting one standard error.}
% \label{fig:iplen}
% \end{figure}
{\bf Dataset Ablation.}
\rev{Next we showcase the need for synthetic data. Intuitively, the majority of our real datasets have commonly found granularities like hourly, daily etc which have specific periodic patterns like 24 time-point period for hourly data. This can make the model not generalize well to underrepresented frequencies. We train a 200M model with no synthetic data added in the mix and showcase the performance on Monash and ETT datasets in Figure~\ref{fig:datasets-abl}. It can be seen that there is a performance drop on Monash because many of the datasets in Monash have under-represented granularities like quarterly, yearly or 10 minutes etc. Perhaps even more compelling is the comparison on ETT datasets. We can see that there is almost no difference between the two models on the hourly ETTh datasets which has a well represented granularity. However, for the 15min ETTm datasets the model with synthetic data performs quite a bit better.}
% \begin{figure}[!ht]
%     \centering
%     \includegraphics[width=0.8\linewidth]{dataset_abl_new.png}
% \caption{Average scaled MAE for Monash on the left and average MAE on ETT datasets on the right. We compare the performance of 200M model with and without the synthetic data.}
% \label{fig:datasets-abl}
% \end{figure}

\rev{We provide a finetuning study in the same setting as~\citep{zhou2023one} in Appendix~\ref{sec:finetuning}, where our model performs better than all baselines on all the reported datasets. This shows the utility of our model on downstream tasks.}

\section{Conclusion}
In this paper, we presented \ours,  a practical foundation model for forecasting whose zero-shot performance comes close to the accuracy of fully-supervised forecasting models on a diverse set of time-series data. This model is pretrained on real-world and synthetic datasets comprising O(100B) timepoints. \rev{We discuss limitations and future work in more detail in Appendix~\ref{sec:lims}}.

\section{Impact Statement}
\rev{This paper shows that it is possible to train a single pretrained model that has phenomenal zero-shot performance on a variety of forecasting tasks, thus opening up exciting possibilities for downstream applications. Therefore it is crucial to discuss ethical and societal considerations of using such a model and how some of the related concerns can be mitigated.}

{\it Data Privacy.} \rev{Note that most of our data sources are publicly available and are aggregated i.e no individual user activity constitutes a time-point. Further the Google Trends data is differentially private.}

{\it Bias.} \rev{Biases can creep into a foundation model through a variety of sources especially through data. The model might perpetuate these biases in forecasts, leading to unfair outcomes. Biased forecasts can have real-world consequences. For example, a biased forecast of crime rates in a neighborhood could lead to increased police presence, disproportionately impacting certain communities. Note that our model is not trained with any covariates so some of these sensitivities are reduced but cannot be ruled out.}

\rev{In this regard, we believe that it is best to release the exact details of the datasets used for training and therefore we have summarized our data sources in Table~\ref{tab:train_data}. Moreover we plan to do a open weights release of our model so that the community can analyze the model for downstream tasks. We will ensure that we release a good model card~\citep{mitchell2019model}. This will also aid in finetuning the model with more diverse data-sources.}

{\it Training cost.} \rev{Our largest model has 200M parameters which is much smaller compared to even smaller scale LLMs. Our data volume is quite large but still smaller compared to SOTA LLMs. We have revealed the exact computation requirements for the models i.e 16 core TPUv5e for 2 days. Note that experimentation and trial runs of course cost more than the final model run. Therefore in the interest of equitability we would like to release the weights of our model in a responsible manner.}

\rev{Lastly we would like to note that similar to LLMs there could be inputs on which the model does not perform well or hallucinates. Therefore in many critical use cases it might be recommended to use the model in a human-in-the-loop fashion or alternatively do a wide range of testing or finetuning.}

\bibliography{refs}
\bibliographystyle{alpha}
\clearpage

\onecolumn 
\appendix
\section{Appendix}

\subsection{Limitations and Future Work}
\label{sec:lims}

\rev{Our work shows that we can train a 200M parameter pretrained forecasting model that has impressive zero-shot performance on a variety of real world forecasting benchmarks with different context and horizon lengths. In this section we would like to discuss limitations and future work.}

{\it Prompt Tuning.} \rev{In LLMs it is well known that prompt tuning techniques like chain-of-thought~\citep{wei2022chain} can drastically improve performance in cases where the model is inaccurate with simple prompts. Such techniques are less clear for time-series foundation model. We can tune simple hyper-parameters like context length as the moment. However, with probabilistic forecasting we might be able to output different statistics as well as come up with techniques that align more with user's expectations while not decreasing likelihood.}

{\it Probabilistic Forecasting.} \rev{It should be straightforward to train with probabilistic loss functions in our framework as detailed in the "Loss Function" part of Section~\ref{sec:model}. However, being one of the first works of building a single foundation model for forecasting, this was not our main focus and is left to future explorations. Note that as mentioned before we plan to release our model weights and after that such loss functions~\citep{salinas2020deepar, awasthi2021benefits} can be added during finetuning.}

{\it Covariate handling.} \rev{Currently the model is not pretrained with covariates as one of the key challenges is finding large volumes of pretrained data with meaningful covariates (apart from date features). We also need methods to have a joint universal representation of covariates. Currently there are two simple techniques we can think of for handling covariates (i) In a zero-shot setting at inference time we can predict in-context and linearly regress the residual on covaraites. Then our model + the residual model can be used for forecasting in the horizon. (ii) during finetuning it is straightforward to handle covariates by adding them as inputs to the input and output residual blocks. Categorical variables can be added as embeddings.}

{\it More finetuning studies.} \rev{We perform a fintuning study in Appendix~\ref{sec:finetuning} following a prior work. However, a more in depth study that involves finetuning in the presence of covariates would be beneficial. This being one of the first works of building a single foundation model for forecasting, this was not our main focus and is left to future explorations. Ideas in recent work such as~\citep{chen2023tsmixer} could be useful in this regard.}

{\it Other architectures.} \rev{Given the cost of training foundation models we did not perform much hyper-parameter tuning in our pretraining, while following some well established best practices for training transformers. In a similar vein, it would also be interesting to try out alternatives like the exciting directions of all MLP structures like~\citep{chen2023tsmixer} or efficient linear state space models like Mamba~\citep{gu2023mamba} (and references there in).}

{\it Interpretability.} \rev{Deep foundation models trained on a huge corpuses of data could be inherently less interpretable compared to statistical methods like ARIMA, ETS~\citep{box1968some}. In this regard methods like LOCO, SHAP (see~\citep{verdinelli2023feature} and references there in) could be used to some extent to attribute feature importances to different lags in the context supplied to the model. However, this does not solve the problem to a full extent and one of the best things to do would be to open source a version of the model with a proper model card.~\citep{mitchell2019model}.}

\subsection{Metrics}
\label{app:metrics}
The metrics that are used for reporting results in this paper are:

\begin{itemize}
    \item MAE~\citep{godahewa2021monash}
    \begin{equation}
    \mathrm{MAE} \left(\*y_{L+1:L+H},  \hat{\*y}_{L+1:L+H}\right) \\
    = \frac{1}{H} \norm{\*y_{L+1:L+H} - \hat{\*y}_{L+1:L+H}}_1. \numberthis \label{eq:mae}
    \end{equation}
    \item msMAPE~\citep{godahewa2021monash}
    \begin{equation}
     \mathrm{msMAPE} \left(\*y_{L+1:L+H},  \hat{\*y}_{L+1:L+H}\right) \\
    = \frac{1}{H} \sum_{i=1}^{H} \frac{2\lvert y_{L+i} - \hat{y}_{L+i}\rvert }{\max\left\{\lvert y_{L+i} \rvert + \lvert \hat{y}_{L+i}\rvert + \epsilon , 0.5 + \epsilon \right\}}. \numberthis \label{eq:msmape}
    \end{equation}
    In Monash benchmarks~\citep{godahewa2021monash} $\epsilon=0.1$ was used. This metric is used in order to avoid undefined values in other normalized metrics like MAPE. In multivariate datasets the metrics are calculated for each time-series and then we take the mean or the median. In this paper we only use the mean versions.

\end{itemize}

{\bf Aggregating across datasets.} Since the datasets have wildly different scales averaging unnormalized metrics like MAE is not kosher. Therefore following~\citep{gruver2023large} we scale the metric of each baseline for a dataset by the same metric achieved by a naive baseline on that dataset. The naive baseline just makes the constant prediction $y_{L}$ repeated across the prediction length. We did not need to do that for the Informer datasets since on these datasets metrics are usually reported on standard normalized data~\citep{nie2022time}.

\subsection{Finetuning study on ETT}
\label{sec:finetuning}
\rev{In this section, we test whether \ours~can be fintuned on a small fraction of a dataset to provide even better performance. We follow the same protocol as in GPT4TS~\citep{zhou2023one} (see Table 13 in their paper).  \citep{zhou2023one} finetune GPT2 input and output blocks on long-term forecasting benchmarks on 10\% of the of the original datasets and compare it against models trained from scratch on the same data. Then the models are evaluated on the original test set task of~\citep{zhou2021informer}. We also tune the input and output residual blocks on 10\% of the training set and present the results in Table~\ref{tab:finetune}. We can see that our model performs the best by a large margin. In ETTh1, ETTh2, ETTm1 our finetuned model is better than 18\%, 3\% and 12\% better than GPT4TS, respectively. In fact we can see our 10\% finetuned model's performances are comparable or better than that of most baselines trained on the whole training dataset as reported in Table 14 of ~\citep{zhou2023one}. This shows that the inductive biases encoded in our model weights by finetuning on a large time-series corpus are better for downstream forecasting task than an off the shelf language model like GPT2, even though our model is orders of magnitude smaller.} 

\begin{table}[h]
\centering
\caption{\rev{MAE of different methods on ETT datasets. All methods use 10\% of the original training set for training or finetuning. The baseline numbers are from Table 13 in~\citep{zhou2021informer}}.}
\label{tab:finetune}
\resizebox{0.9\textwidth}{!}{%
\begin{tabular}{l|l|c|c|c|c|c|c|c}
\toprule
\multicolumn{2}{c|}{Dataset} & \ours(FT) & GPT4TS(FT) & DLinear & PatchTST & TimeNet & FEDFormer & Autoformer  \\
\midrule
\multirow{5}{*}{ETTh1} & 96 & 0.398 & 0.456 & 0.495 & 0.485 & 0.628 & 0.499 & 0.552    \\
                      & 192 & 0.424 & 0.516 & 0.538 & 0.524 & 0.593 & 0.555 & 0.598    \\
                      & 336 & 0.436 & 0.535 & 0.622 & 0.550 & 0.648 & 0.574 & 0.619  \\
                      & 720 & 0.445 & 0.591 & 0.743 & 0.610 & 0.641 & 0.614 & 0.616   \\
                      & Avg & \textbf{0.426} & 0.525 & 0.600 & 0.542 & 0.628 & 0.561 & 0.596  \\
\midrule
\multirow{5}{*}{ETTh2} & 96 & 0.356 & 0.374 & 0.411 & 0.389 & 0.409 & 0.416 & 0.451 \\
                      & 192 & 0.400 & 0.411 & 0.519 & 0.414 & 0.467 & 0.474 & 0.477 \\
                      & 336 & 0.428 & 0.433 & 0.572 & 0.441 & 0.494 & 0.501 & 0.543 \\
                      & 720 & 0.457 & 0.464 & 0.648 & 0.480 & 0.491 & 0.509 & 0.523  \\
                      & Avg & \textbf{0.410} & 0.421 & 0.538 & 0.431 & 0.465 & 0.475 & 0.499  \\
\midrule
\multirow{5}{*}{ETTm1} & 96 & 0.345 & 0.404 & 0.392 & 0.419 & 0.501 & 0.518 & 0.614 \\
                      & 192 & 0.374 & 0.423 & 0.412 & 0.434 & 0.528 & 0.546 & 0.592  \\
                      & 336 & 0.397 & 0.439 & 0.434 & 0.454 & 0.568 & 0.775 & 0.677 \\
                      & 720 & 0.436 & 0.498 & 0.477 & 0.556 & 0.549 & 0.579 & 0.630 \\
                      & Avg & \textbf{0.388} & 0.441 & 0.429 & 0.466 & 0.537 & 0.605 & 0.628  \\
\midrule
\multirow{5}{*}{ETTm2} & 96 & 0.263 & 0.269 & 0.303 & 0.274 & 0.285 & 0.399 & 0.454  \\
                      & 192 & 0.309 & 0.309 & 0.345 & 0.317 & 0.323 & 0.379 & 0.691  \\
                      & 336 & 0.349 & 0.346 & 0.385 & 0.353 & 0.353 & 0.559 & 1.407  \\
                      & 720 & 0.415 & 0.417 & 0.440 & 0.427 & 0.449 & 0.614 & 1.166  \\
                      & Avg & \textbf{0.334} & 0.335 & 0.368 & 0.343 & 0.353 & 0.488 & 0.930  \\
\midrule
\end{tabular}
}
\end{table}

\subsection{Pretraining PatchTST}
\label{app:pretrain_patchtst}
\rev{Since \ours~applies a similar patching strategy as PatchTST~\citep{nie2022time}, for an ablation study we use the same data loader and pretrain a PatchTST model of 200M parameters to the same number of FLOPS as the final 200M \ours~model. We denote it as PatchTST(ZS). The two models share the same hyperparameters of the transformer stack. For PatchTST(ZS) we use the same input patch length = 32, and a stride of length half of input patch size (i.e. stride = 16) as done in the original PatchTST paper.}

\rev{We report the detailed results on Monash and ETT in Appendix~\ref{app:monash} and~\ref{app:informer}. It can be seen that the results are not that good for PatchTST(ZS) on Monash. This is expected since our pretrain data loader will predominantly have context lengths of 512 instead of shorter context lengths as in Monash. Moreover the PatchTST model does fewer iterations at the same number of FLOPS. On ETT datasets, the PatchTST(ZS) model is performs similarly to TimesFM(ZS) and PatchTST. This is also expected since the context length for this study is indeed 512.}

\rev{As PatchTST(ZS) is an encoder-decoder model, to pretrain it for zero shot forecasting one should theoretically prepare all possible context lengths and horizon lengths in the pretrain datasets. Pretraining it to its maximum performance requires much more compute and likely more careful tuning compared to pretraining \ours.}
% Our metrics on PatchTST(ZS) here should be close to, whereas do not reflect, its full potential as a pretrained foundation model.}

\subsection{Additional Empirical Results} 
\label{app:moreexp}
\begin{figure*}[!ht]
  \centering
  \begin{subfigure}[b]{0.3\textwidth}
     \includegraphics[width=\textwidth]{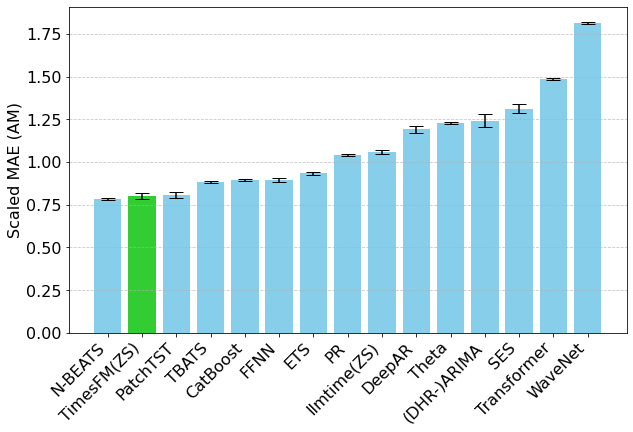} % Replace 'placeholder1' with your image filename
     \caption{\scriptsize{\rev{Monash Archive}~\citep{godahewa2021monash}}}
     \label{fig:monash_mae_am}
  \end{subfigure}
  \hfill % Add some space between subfigures
  \begin{subfigure}[b]{0.3\textwidth}
     \includegraphics[width=\textwidth]{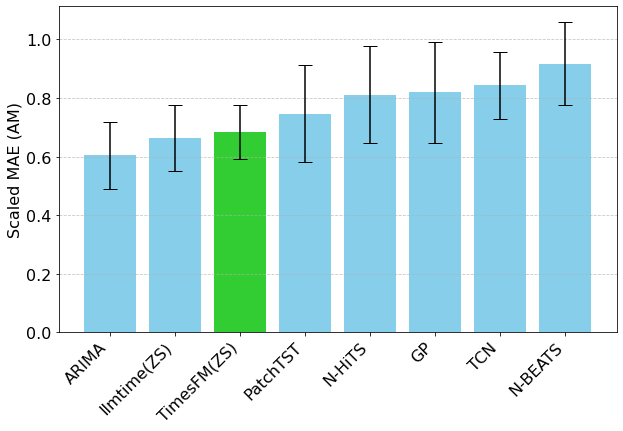} % Replace 'placeholder2' with your image filename
     \caption{\scriptsize{\rev{Darts}~\citep{herzen2022darts}}}
     \label{fig:darts_am}
  \end{subfigure}
  \hfill % Add some space between subfigures
  \begin{subfigure}[b]{0.3\textwidth}
     \includegraphics[width=\textwidth]{autoformer_chart_200m_final.png} % Replace 'placeholder3' with your image filename
     \caption{\scriptsize{ETT (Horizons 96 and 192)~\citep{zhou2021informer}}}
     \label{fig:informer_am}
  \end{subfigure}
\caption{\rev{We report average performance in three groups of datasets. In all figures, the lower the metric the better and the error bars represent one standard error. Note that among the baselines only \ours~and llmtime are zero-shot. In (a) we report results on the Monash datasets. \rev{Since the datasets have different scales, we take the Arithmetric Mean (AM) the MAE's scaled by the MAE of a naive baseline}. We can see that \ours~is within significance of the top model N-BEATS. In (b), we report the similarly scaled MAE on the Darts benchmarks. \ours~is within significance of the top of method which is ARIMA in this case. Note that these datasets have one time-series each and therefore statistical methods are competitive with deep learning ones. Finally, in (c) we report the average MAE for 96 and 192 horizon prediction tasks on 4 ETT datasets i.e 8 tasks in total. \ours~and PatchTST are the best performing models in this case.}}
  \label{fig:all_bar_charts_am}
\end{figure*}

In this section, we provide more detailed tables for our zero-shot datasets and experiments described in Section~\ref{sec:zeroshot}. The AM based aggregated metrics are presented in Figure~\ref{fig:all_bar_charts_am}.

\subsubsection{Darts}
\label{app:darts}
We present the MAE results individually from all 8 datasets in Table~\ref{tab:darts}. It can be seen that \ours~performs well for all datasets with clear seasonal patterns. On an average we are within significant level of the best model. Note that there are only 8 time-series as a whole in Darts and theerfore these evaluations have very wide confidence intervals.

In Figure~\ref{fig:darts_all_4x4} we present visual comparisons of our forecasts vs some of the baselines.
\begin{table}[!ht]
\centering
\caption{MAE for Darts datasets. We also include the naive baseline that predicts the last values in the context repeatedly.}
\label{tab:darts}
\resizebox{\textwidth}{!}{
\begin{tabular}{lrrrrrrrrr}
\toprule
 & GP & ARIMA & TCN & N-BEATS & N-HiTS & llmtime(ZS) & \ours(ZS) & \rev{PatchTST} & NAIVE \\
\midrule
AirPassengersDataset & 34.67 & 24.03 & 54.96 & 97.89 & 59.16 & 34.37 & \rev{62.51} & \rev{44.65} & 81.45 \\
AusBeerDataset & 102.05 & 17.13 & 30.90 & 10.39 & 34.23 & 16.13 & \rev{11.94} & \rev{21.97} & 96.35 \\
GasRateCO2Dataset & 2.27 & 2.37 & 2.64 & 2.63 & 3.85 & 3.50 & \rev{2.50} & \rev{2.67} & 2.29 \\
MonthlyMilkDataset & 30.33 & 37.19 & 70.86 & 33.64 & 32.73 & 9.68 & \rev{28.09} & \rev{42.60} & 85.71 \\
SunspotsDataset & 53.74 & 43.56 & 51.82 & 73.15 & 49.93 & 47.34 & \rev{41.40} & \rev{62.33} & 48.24 \\
WineDataset & 4552.06 & 2306.70 & 3287.14 & 4562.02 & 3909.51 & 1569.32 & \rev{2871.33} & \rev{2498.69} & 4075.28 \\
WoolyDataset & 649.98 & 588.78 & 1158.79 & 903.01 & 382.09 & 808.73 & \rev{728.92} & \rev{542.28}  & 1210.33 \\
HeartRateDataset & 5.65 & 5.56 & 5.49 & 6.57 & 6.10 & 6.21 & \rev{5.85} & \rev{6.74} & 5.92 \\
\midrule
Scaled MAE (Arithmetic Mean) & 0.8193 & 0.6045 & 0.8427 & 0.9176 & 0.8109 & 0.6641 & 0.6829 & \rev{0.7462} &1.0000 \\
Scaled MAE (Geometric Mean) & 0.7509 & 0.5219 & 0.7946 & 0.7316 & 0.6936 & 0.4882 & 0.5767 & \rev{0.6458} &1.0000 \\
\bottomrule
\end{tabular}
}
\end{table}

\subsubsection{Monash}
\label{app:monash}

\begin{table}[!ht]
\centering
\caption{We present the mean MAE results for our methods along size Monash baselines. We also include the naive baseline that predicts the last values in the context repeatedly.}
\label{tab:monash_mae}
\resizebox{1\textwidth}{!}{
\begin{tabular}{lrrrrrrrrrrrrrrrr}
\toprule
 & llmtime(ZS) & SES & Theta & TBATS & ETS & (DHR-)ARIMA & PR & CatBoost & FFNN & DeepAR & N-BEATS & WaveNet & Transformer & \rev{PatchTST(ZS)} & \ours(ZS) &  NAIVE \\
Dataset &  &  &  &  &  &  &  &  &  &  &  &  &  &  &  &  \\
\midrule
australian electricity demand & 459.96 & 659.60 & 665.04 & 370.74 & 1282.99 & 1045.92 & 247.18 & 241.77 & 258.76 & 302.41 & 213.83 & 227.50 & 231.45 & \rev{382.23} & \rev{448.81} & 659.60 \\
bitcoin & 1.75e18 & 5.33e18 & 5.33e18 & 9.90e17 & 1.10e18 & 3.62e18 & 6.66e17 & 1.93e18 & 1.45e18 & 1.95e18 & 1.06e18 & 2.46e18 & 2.61e18 & \rev{1.11e18} & \rev{1.3e18} & \rev{7.77e17} \\
pedestrian counts & 70.20 & 170.87 & 170.94 & 222.38 & 216.50 & 635.16 & 44.18 & 43.41 & 46.41 & 44.78 & 66.84 & 46.46 & 47.29 & \rev{51.27} & \rev{40.71} & 170.88 \\
weather & 2.32 & 2.24 & 2.51 & 2.30 & 2.35 & 2.45 & 8.17 & 2.51 & 2.09 & 2.02 & 2.34 & 2.29 & 2.03 & \rev{2.07} & \rev{2.07} & 2.36 \\
nn5 daily & 9.39 & 6.63 & 3.80 & 3.70 & 3.72 & 4.41 & 5.47 & 4.22 & 4.06 & 3.94 & 4.92 & 3.97 & 4.16 & \rev{3.77} & \rev{3.54} & 8.26 \\
nn5 weekly & 15.91 & 15.66 & 15.30 & 14.98 & 15.70 & 15.38 & 14.94 & 15.29 & 15.02 & 14.69 & 14.19 & 19.34 & 20.34 & \rev{17.00} & \rev{14.67} & 16.71 \\
tourism yearly & 140081.78 & 95579.23 & 90653.60 & 94121.08 & 94818.89 & 95033.24 & 82682.97 & 79567.22 & 79593.22 & 71471.29 & 70951.80 & 69905.47 & 74316.52 & \rev{224411.89} & \rev{109977.29} & 99456.05 \\
tourism quarterly & 14121.09 & 15014.19 & 7656.49 & 9972.42 & 8925.52 & 10475.47 & 9092.58 & 10267.97 & 8981.04 & 9511.37 & 8640.56 & 9137.12 & 9521.67 & \rev{21276.98} &  \rev{12102.04} & 15845.10 \\
tourism monthly & 4724.94 & 5302.10 & 2069.96 & 2940.08 & 2004.51 & 2536.77 & 2187.28 & 2537.04 & 2022.21 & 1871.69 & 2003.02 & 2095.13 & 2146.98 & \rev{4596.21} & \rev{3183.77} & 5636.83 \\
cif 2016 & 715086.33 & 581875.97 & 714818.58 & 855578.40 & 642421.42 & 469059.49 & 563205.57 & 603551.30 & 1495923.44 & 3200418.00 & 679034.80 & 5998224.62 & 4057973.04 & \rev{8374813.14} & \rev{773980.44} & 386526.37 \\
covid deaths & 304.68 & 353.71 & 321.32 & 96.29 & 85.59 & 85.77 & 347.98 & 475.15 & 144.14 & 201.98 & 158.81 & 1049.48 & 408.66 & \rev{348.60} & \rev{209.80} & \rev{353.71} \\
fred md & 2013.49 & 2798.22 & 3492.84 & 1989.97 & 2041.42 & 2957.11 & 8921.94 & 2475.68 & 2339.57 & 4264.36 & 2557.80 & 2508.40 & 4666.04 & \rev{4965.51} & \rev{947.12} & 2825.67 \\
traffic hourly & 0.03 & 0.03 & 0.03 & 0.04 & 0.03 & 0.04 & 0.02 & 0.02 & 0.01 & 0.01 & 0.02 & 0.02 & 0.01 & \rev{0.01} & \rev{0.01} & 0.03 \\
traffic weekly & 1.17 & 1.12 & 1.13 & 1.17 & 1.14 & 1.22 & 1.13 & 1.17 & 1.15 & 1.18 & 1.11 & 1.20 & 1.42 & \rev{1.23} & \rev{1.12} & 1.19 \\
saugeenday & 28.63 & 21.50 & 21.49 & 22.26 & 30.69 & 22.38 & 25.24 & 21.28 & 22.98 & 23.51 & 27.92 & 22.17 & 28.06 & \rev{22.33} & \rev{24.6}3 & 21.50 \\
us births & 459.43 & 1192.20 & 586.93 & 399.00 & 419.73 & 526.33 & 574.93 & 441.70 & 557.87 & 424.93 & 422.00 & 504.40 & 452.87 & \rev{1193.28} & \rev{437.27} & 1152.67 \\
hospital & 24.62 & 21.76 & 18.54 & 17.43 & 17.97 & 19.60 & 19.24 & 19.17 & 22.86 & 18.25 & 20.18 & 19.35 & 36.19 & \rev{20.87} & \rev{19.41} & 24.07 \\
solar weekly & 2049.09 & 1202.39 & 1210.83 & 908.65 & 1131.01 & 839.88 & 1044.98 & 1513.49 & 1050.84 & 721.59 & 1172.64 & 1996.89 & 576.35 & \rev{1093.46} & \rev{1258.27} & 1729.41 \\
\midrule 
Scaled MAE (Arithmetic Mean) & 1.0588 & 1.3115 & 1.2295 & 0.8824 & 0.9337 & 1.2427 & 1.0382 & 0.8924 & 0.8942 & 1.1925 & 0.7844 & 1.8119 & 1.4840 & 2.1419 & 0.8005 & 1.0000 \\
Scaled MAE (Geometric Mean) & 0.9715 & 1.0855 & 0.9371 & 0.7736 & 0.8104 & 0.9449 & 0.8218 & 0.7733 & 0.7044 & 0.7477 & 0.7005 & 0.9384 & 0.8619 & 1.0557 & 0.6846 & 1.0000 \\
\bottomrule
\end{tabular}
}

\end{table}

In Table~\ref{tab:monash_mae} we present the actual MAE numbers that are behind the main Figure~\ref{fig:monash_mae}. In Figure~\ref{fig:monash_4x4}, we present some examples of our zero-shot forecasts. For most datasets, we set the context window to be the maximum length of the series in the dataset capped at 512 (similar to statistcal models used in the official Monash baselines). \rev{For some datasets, we did some inference time-tuning of the context length, i.e we predict the last horizon length number of points in the training set with context lengths 32, 64 and maximum allowed and chose the best one in terms of this validation metric. This is fair as most Monash DL baselines use different context lengths for different datasets during training and our model is completely zero-shot. The max context lengths used for these datasets are (cif 2016, 32), (tourism yearly, 32), (covid deaths, 32), (bitcoin, 32), (tourism monthly, 32) and (tourism monthly, 64)}.

\subsubsection{Informer}
\label{app:informer}
We present the MAE on the last split of the test set for all dataset, horizon pairs considered in Table~\ref{tab:inf}. Owing to expensive evaluations for llmtime, the results are reported on the last test window of the original test split, as done in~\citep{gruver2023large}.

\begin{table}[!ht]
\centering
\caption{MAE for ETT datasets for prediction horizons 96 and 192. Owing to expensive evaluations for llmtime, the results are reported on the last test window of the original test split.}
\label{tab:inf}
\resizebox{1\textwidth}{!}{
\begin{tabular}{lrrrrrrr}
\toprule
& llmtime(ZS)* & PatchTST & \rev{PatchTST(ZS)} & FEDFormer & AutoFormer & Informer  & TimesFM(ZS) \\
Dataset &  &  &  &  &  &  & \\
\midrule
ETTh1 (horizon=96)   & 0.42 & 0.41 & \rev{0.39} & 0.58 & 0.55 & 0.76 & \rev{0.45} \\
ETTh1 (horizon=192)  & 0.50 & 0.49 & \rev{0.50} & 0.64 & 0.64 & 0.78 & \rev{0.53} \\
ETTh2 (horizon=96)   & 0.33 & 0.28 & \rev{0.37} & 0.67 & 0.65 & 1.94 & \rev{0.35} \\
ETTh2 (horizon=192)  & 0.70 & 0.68 & \rev{0.59} & 0.82 & 0.82 & 2.02 & \rev{0.62} \\
ETTm1 (horizon=96)   & 0.37 & 0.33 & \rev{0.24} & 0.41 & 0.54 & 0.71 & \rev{0.19} \\
ETTm1 (horizon=192)  & 0.71 & 0.31 & \rev{0.26} & 0.49 & 0.46 & 0.68 & \rev{0.26} \\
ETTm2 (horizon=96)   & 0.29 & 0.23 & \rev{0.22} & 0.36 & 0.29 & 0.48 & \rev{0.24} \\
ETTm2 (horizon=192)  & 0.31 & 0.25 & \rev{0.22} & 0.25 & 0.30 & 0.51 & \rev{0.27} \\
\midrule
Avg                  & 0.45 & 0.37 & 0.35        & 0.53 & 0.53 & 0.99 & 0.36 \\ 
\bottomrule
\end{tabular}
}
\end{table}

% \begin{table}[!ht]
% \centering
% \caption{MAE for ETT datasets for prediction horizons 96 and 192. Owing to expensive evaluations for llmtime, the results are reported on the last test window of the original test split.}
% \label{tab:inf_alt}
% \begin{tabular}{lrrrrrr}
% \toprule
%  & llmtime(ZS) & PatchTST & \ours(ZS) & FEDFormer & AutoFormer & Informer \\
% Dataset &  &  &  &  &  &  \\
% \midrule
% ETTh1 (Horizon=96) & 0.42 & 0.41 & 0.37 & 0.58 & 0.55 & 0.76 \\
% ETTh1 (Horizon=192) & 0.50 & 0.49 & 0.49 & 0.64 & 0.64 & 0.78 \\
% ETTh2 (Horizon=96) & 0.33 & 0.28 & 0.28 & 0.67 & 0.65 & 1.94 \\
% ETTh2 (Horizon=192) & 0.70 & 0.68 & 0.58 & 0.82 & 0.82 & 2.02 \\
% ETTm1 (Horizon=96) & 0.37 & 0.33 & 0.25 & 0.41 & 0.54 & 0.71 \\
% ETTm1 (Horizon=192) & 0.71 & 0.31 & 0.24 & 0.49 & 0.46 & 0.68 \\
% ETTm2 (Horizon=96) & 0.29 & 0.23 & 0.28 & 0.36 & 0.29 & 0.48 \\
% ETTm2 (Horizon=192) & 0.31 & 0.25 & 0.24 & 0.25 & 0.30 & 0.51 \\
% \bottomrule
% \end{tabular}
% \end{table}
\subsection{More Details on Models}
\label{app:moremodels}
We now present implementation details about \ours~and other baselines.

{\bf \ours.} For our main 200M model we use 16 attention heads, 20 layers, a input patch length of 32 and output patch length of 128. The model dimension is set to 1280. We train with layer norm and a cosine decay learning rate schedule with peak learning rate of $5e-4$. The hyper-parameters of \ours~for various sizes are provided in Table~\ref{tab:hparams}. Note that the settings are for the base models and not ablation models. The hidden dims of both the residual block and the FFN in the transformer layers are set as the same as model dimensions. We keep layer norm in transformer layers but not in the residual blocks.

\begin{table}[!ht]
\centering
\caption{Hyper-parameters for \ours}
\label{tab:hparams}
\begin{tabular}{lrrrrrr}
\toprule
 & num\textunderscore layers & model\textunderscore dims & output\textunderscore patch\textunderscore len& input\textunderscore patch\textunderscore len & num\textunderscore heads & dropout \\
Size &  &  &  &  &  &  \\
\midrule
200M  & 20 & 1280 & 128 & 32 & 16 & 0.2 \\
70M  & 10 & 1024 & 128 & 32 & 16 & 0.2 \\
17M  & 10 & 512 & 128 & 32 & 16 & 0.2 \\
\bottomrule
\end{tabular}
\end{table}

{\bf Monash Baselines.} The raw metrics for the Monash baselines are directly taken from Tables 9 and 11 of the supplementary material of the original paper~\citep{godahewa2021monash}. For llmtime, we use the precomputed outputs provided by the authors of ~\citep{gruver2023large}.

{\bf Darts Baselines.} For all the Darts baselines we use the precomputed outputs provided by the authors of ~\citep{gruver2023large}. For more details please see Section C.1 in that paper.

{\bf Informer Baselines.} For FEDFormer~\citep{zhou2022fedformer}, Autoformer~\citep{wu2021autoformer}, Informer~\citep{zhou2021informer} and PatchTST~\citep{nie2022time} we use the original hyperparameters and implementation. The results presented in the main paper are obtained on the last test window of length horizon length as stated in the llmtime~\citep{gruver2023large} paper.

We generate the llmtime predictions using the code provided by the authors~\footnote{\url{https://github.com/ngruver/llmtime/blob/main/experiments/run_monash.py}} but adapted to the ETT datasets. Note that as of January 2024, OpenAI has discontinued access to GPT-3, therefore we had to use the GPT-3.5-Turbo model.

\subsection{Date Features}
\label{app:date_features}
As we mentioned earlier, since we are building a single pre-trained model, we cannot have dataset specific dynamic or static covariates during training time. However, the datetime column is ubiquitous in all time-series data, so we can technically have date derived features like day of the week, month of the year etc processed into a vector at each time-point $t$, denoted by $\*x_{t} \in \-R^{r}$. 

If so, the learning task can be rewritten as
\begin{equation*}
     f: \left( \*y_{1:L}, \*x_{1:L+H} \right) \longrightarrow \hat{\*y}_{L+1:L+H}.
\end{equation*}

There are many options to incorporate these features into the model, one being to directly concatenate them after the time-points in each patch. For this paper we decide to focus on the univariate time-series input, and will investigate this enhancement in the future.

\subsection{Synthetic Data}
\label{app:syn}

We create the synthetic data to reflect common time-series patterns using traditional statistical models. We start with four simple times series patterns:
\begin{itemize}
    \item Piece-wise linear trends (I), where the number of the piece-wise linear components is randomly chosen between 2 and 8.
    \item ARMA$(p, q)$ (II), where $1 \leq p, q \leq 8$ and the corresponding coefficients are generated from either a multivariate Gaussian or a uniform, then normalized.
    \item Seasonal patterns. In particular we create the sine (III) and the cosine (IV) waves of different random periods between 4 and \rev{max context length / 2} time-points and time delays. 
\end{itemize}

We then randomly enable / disable these four components (I) - (IV), generate their time-series of length 2048 respectively, and sum them up using uniformly sampled random weights to create each times series in the synthetic datasets. We also choose to apply the trend multiplicatively 50\% of the times the trend component is chosen.

\subsection{Illustrative Examples} 
\label{app:syn_exp}

We conduct a visual inspection of the forecasts generated by \ours, first on some synthetic examples and then on the benchmark datasets.

\begin{figure}[!ht]
    \centering
    \includegraphics[width=1.0\linewidth]{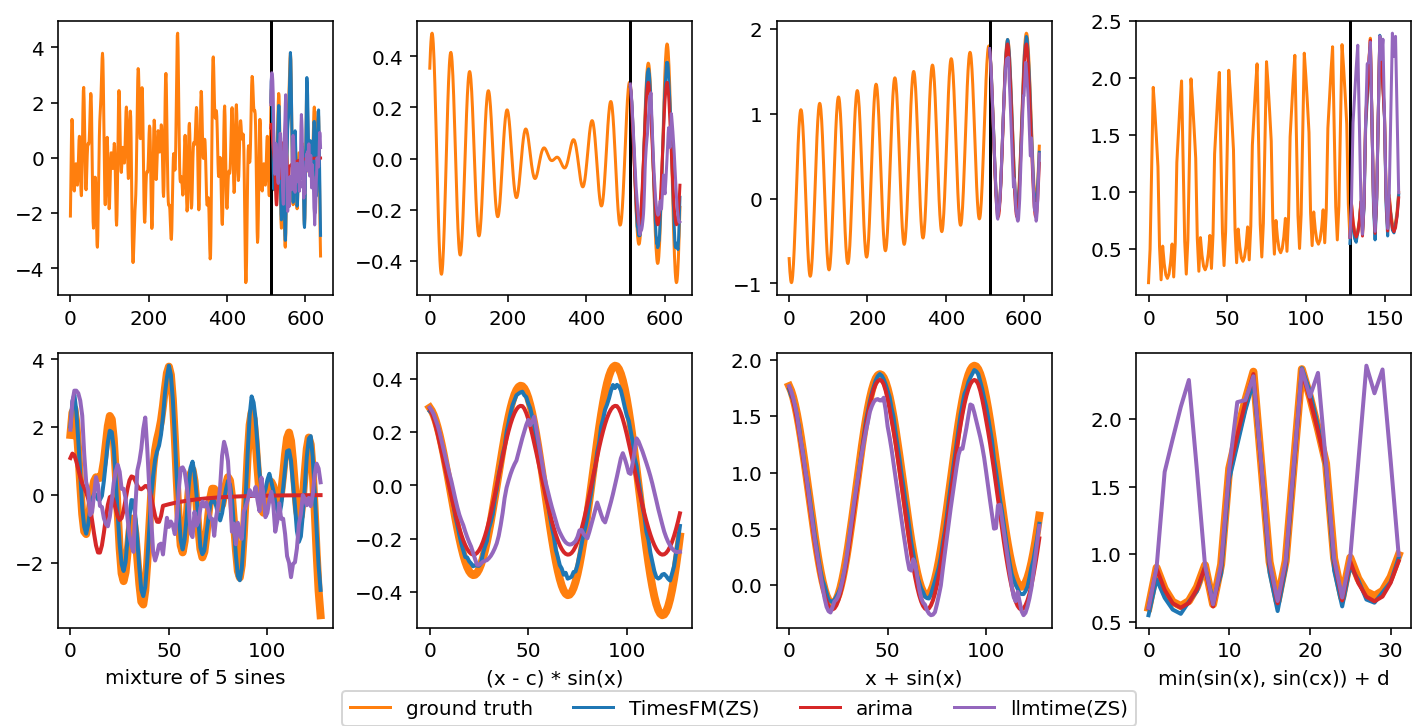}
\caption{Forecasts visualized on synthetic curves. The bottom row plots zoom in on the prediction horizon for the sake of clarity.}
\label{fig:viz-syn-4p}
\end{figure}

In Figure~\ref{fig:viz-syn-4p} we show 4 different synthetic curves: (1) sum of 5 sine curves of different periods, (2) a sine curve linearly scaled, (3) a sine curve with a linear trend, and (4) minimum of two sine curves with a linear trend. Our results suggests that \ours~picks up the trend and seasonal components readily interpretable by humans, while ARIMA and (to a lesser extent) llmtime fail in some of the instances.

\begin{figure}[!ht]
    \centering
    \includegraphics[width=1.0\linewidth]{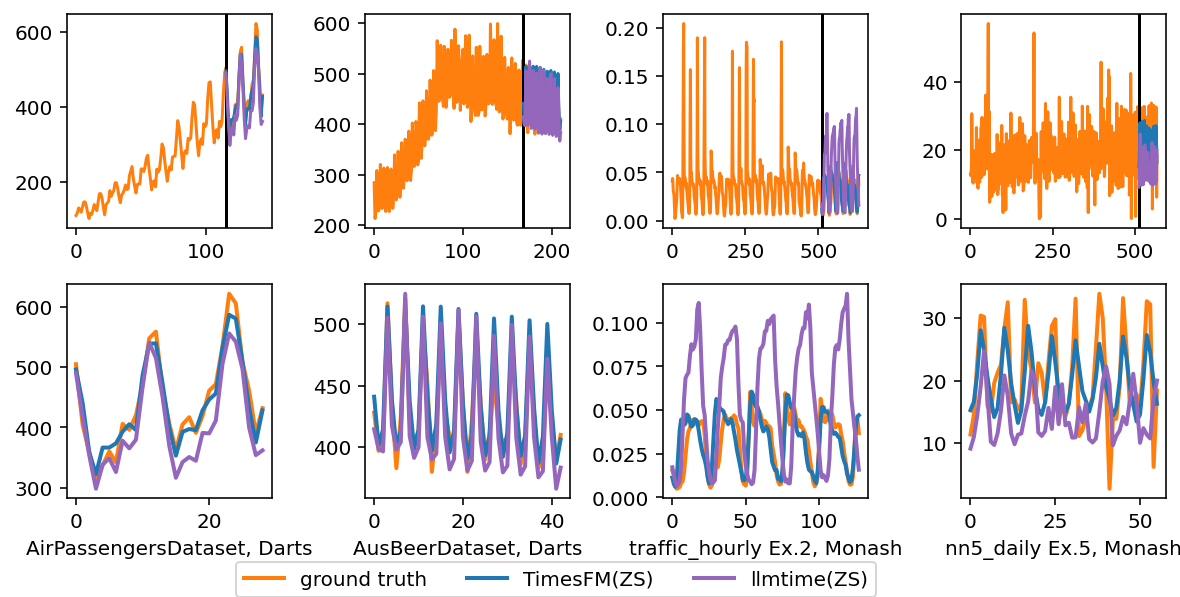}
\caption{Forecasts visualized on Darts and Monash. The bottom row plots zoom in on the prediction horizon for the sake of clarity.}
\label{fig:visualization-4p}
\end{figure}

As illustrated in Figure~\ref{fig:visualization-4p}, \ours~also effectively captures these subtle characteristics within both the trend and seasonal patterns of the depicted real world time-series. For instance, in the Air Passenger dataset, \ours~correctly captures the amplitude increase with trend --this is also reflected by the fact that it attains the best MAE on this dataset (see Table~\ref{tab:darts}). In the traffic hourly example on the left, it can be seen that \ours~can correctly identify the seasonal peaks even in the presence of outliers in the context, while llmtime is thrown off.

\begin{figure}[!ht]
\centering
\includegraphics[width=\linewidth]{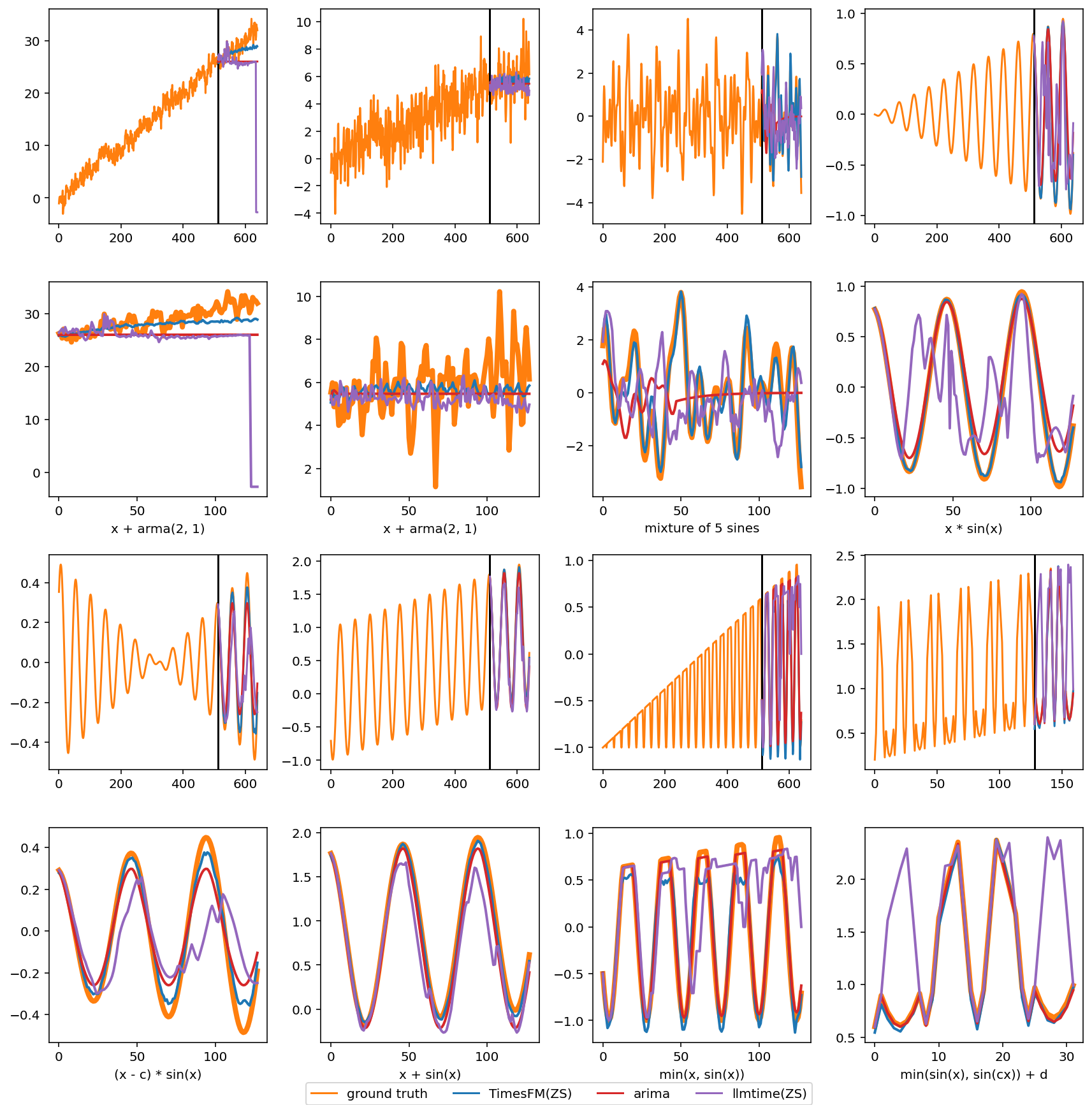}
\caption{Forecasts visualized on synthetic curves. The second row plots zoom in on the prediction horizon for the sake of clarity.}
\label{fig:syn_4x4}
\end{figure}

\begin{figure}[!ht]
\centering
\includegraphics[width=\linewidth]{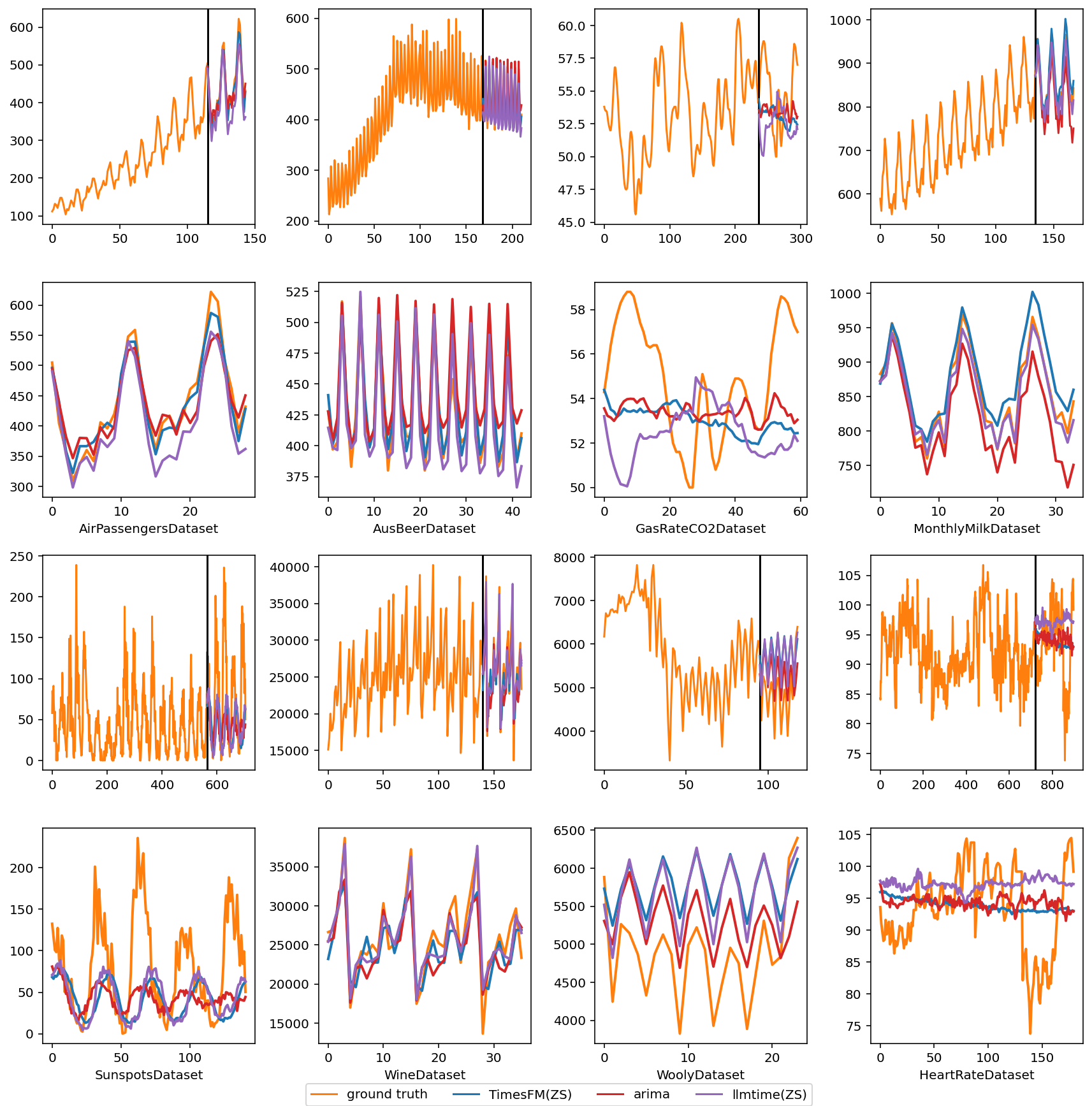}
\caption{Forecasts visualized on all Darts datasets. The second row plots zoom in on the prediction horizon for the sake of clarity.}
\label{fig:darts_all_4x4}
\end{figure}

\begin{figure}[!ht]
\centering
\includegraphics[width=\linewidth]{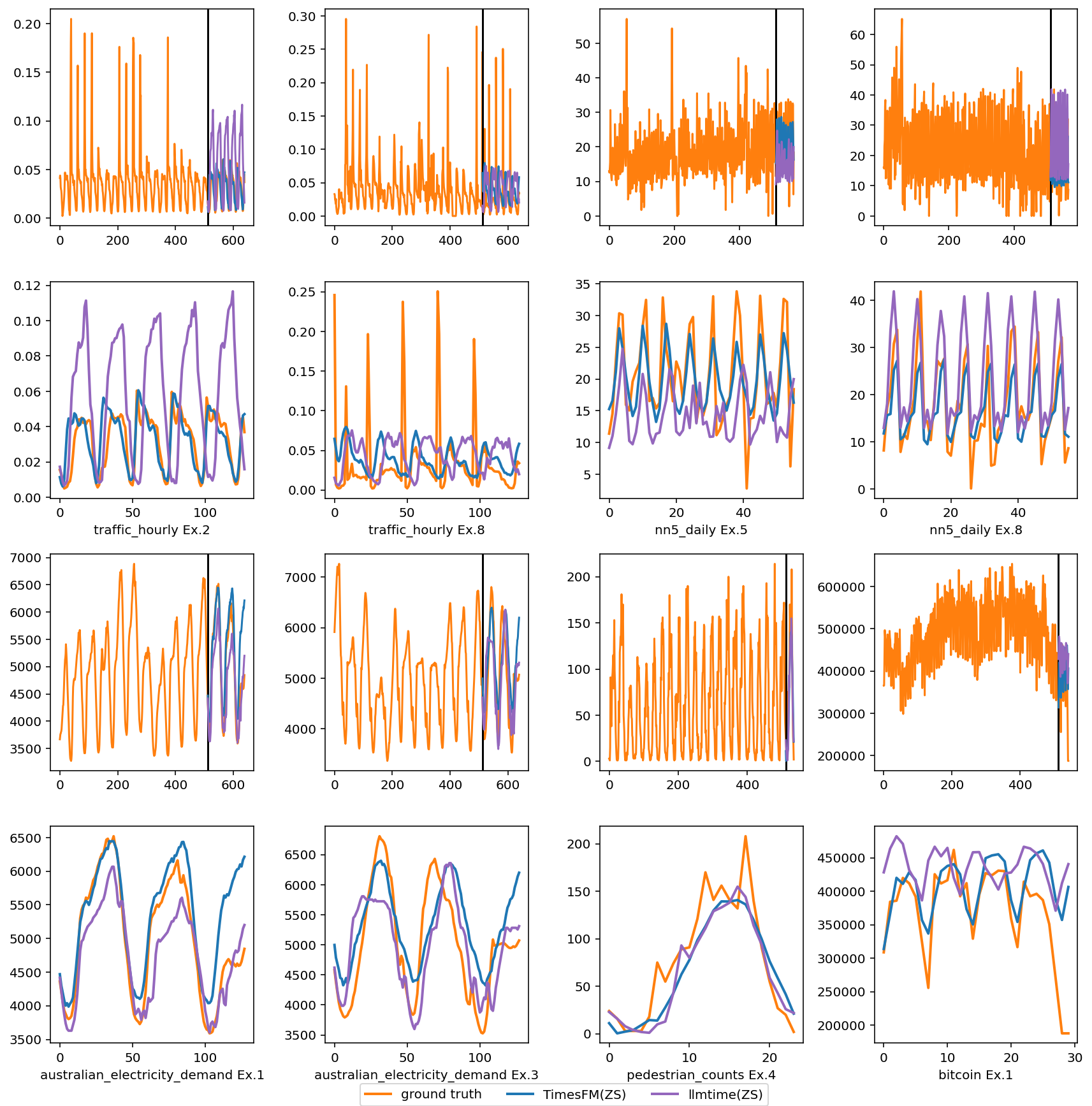}
\caption{Forecasts visualized on a few Monash datasets. The second row plots zoom in on the prediction horizon for the sake of clarity.}
\label{fig:monash_4x4}
\end{figure}

We provide more visualization in Figure~\ref{fig:syn_4x4}, Figure~\ref{fig:darts_all_4x4} and Figure~\ref{fig:monash_4x4}.

% \subsubsection{Synthetic Data Visualizations}

% In this section we aim to investigate how \ours~generalizes to some common temporal patterns that are potentially out of distribution from the synthetic parts of its pretraining dataset. We present some examples in Figure~\ref{fig:syn_4x4}. We also compare how ARIMA and llmtime behaves on these examples.

% In particular, we generate these curves by (i) and (ii) linear trend + ARMA(2, 1), (iii) sum of 5 sines of different periods, (iv) and (v) a sine curve scaled linearly, (vi) a sine curve with a linear trend, (vii) a sine curve capped linearly, and (viii) minimum of two sines with a linear trend.

% We notice that, compared to (Auto)ARIMA and llmtime, \ours~is more capable of following trends and nuanced seasonal patterns. For example, Plot 3 shows the sum of 5 sine curves which is out of the distribution of our synthetic dataset. However \ours~ still correctly identifies the seasonal pattern, likely because similar pattern occurred in the context. Plot 4 and 5 demonstrate that \ours~correctly identifies the multiplicative trend which seems hard for either llmtime or ARIMA.

% It is worth pointing out that \ours~is practically the fastest and the easiest forecasting method to run here. In order for (Auto)ARIMA to work the best one needs to properly identify the seasonal length and the trend. Even so the per time-series optimization takes way longer then \ours~simply decoding.

\end{document}